# DPGIIL: Dirichlet Process-Deep Generative Model-Integrated Incremental Learning for Clustering in Transmissibility-based Online Structural Anomaly Detection

**Lin-Feng Mei[1], Wang-Ji Yan[1,2]***

[1]*State Key Laboratory of Internet of Things for Smart City and Department of Civil and Environmental Engineering, University of Macau, China*

[2]*Guangdong-Hong Kong-Macau Joint Laboratory for Smart Cities, China*

**Abstract:** Clustering based on vibration responses, such as transmissibility functions (TFs), is promising in structural anomaly detection. However, most existing methods struggle to determine the optimal cluster number, handle high-dimensional streaming data, and rely heavily on manually engineered features due to their shallow structures. To address these issues, this work proposes a novel clustering framework, referred to as Dirichlet process-deep generative model-integrated incremental learning (DPGIIL), for online structural anomaly detection, which combines the advantages of deep generative models (DGMs) in representation learning and the Dirichlet process mixture model (DPMM) in identifying distinct patterns in observed data. Within the context of variational Bayesian inference, a lower bound on the log marginal likelihood of DPGIIL, tighter than the evidence lower bound, is derived analytically, which enables the joint optimization of DGM and DPMM parameters, thereby allowing the DPMM to regularize the DGM's feature extraction process. Additionally, a greedy split-merge scheme-based coordinate ascent variational inference method is devised to accelerate the optimization. The summary statistics of the DPMM, along with the network parameters, are used to retain information about previous data for incremental learning. For online structural anomaly detection, DPGIIL can not only detect anomalies by dynamically assigning incoming data to new clusters but also indicate different structural states using distinct clusters, thereby providing additional information about the operating conditions of the monitored structure compared to traditional anomaly detectors. Three case studies demonstrate the dynamic adaptability of the proposed method and show that it outperforms some state-of-the-art approaches in both structural anomaly detection and clustering. The source code is available at: https://github.com/Christine-cmd/DPGIIL.

---

*Corresponding author.
E-mail address: yc17409@um.edu.mo (L.F. Mei); wangjiyan@um.edu.mo (W.J. Yan)

# 1 Introduction

The service life of engineering structures inevitably decreases during long-term operation, which underscores the significance of ensuring their operating safety to prevent unexpected monetary losses and casualties. In recent years, with the rapid development of sensor systems, computing resources, and data processing approaches, vibration-based structural health monitoring (SHM) with machine learning (ML) techniques has become one of the most prevalent methods to ensure structural safety [1-3], which mainly includes unsupervised and supervised methods according to the employed ML techniques. Unsupervised methods are more flexible as they require only training data from normal operating conditions [1]. However, they are mainly used for fundamental tasks, predominantly structural anomaly detection, due to the lack of information about different damage patterns. In contrast, supervised methods can handle higher-level tasks such as damage classification and quantification, but are often limited by the high cost of obtaining well-annotated training data from various damage scenarios [4].

Currently, outlier analysis-based methods have been widely used for structural anomaly detection, which establishes a baseline based on training data from normal conditions, and any deviation from this baseline is identified as a structural anomaly caused by damage [1]. A pivotal milestone in this area was proposed by Worden et al. [5], in which they employed the Mahalanobis squared distance (MSD) of transmissibility functions (TFs) as the damage index and established a statistical threshold based on Monte Carlo simulation to identify structural anomalies. This approach has since been followed and refined in numerous studies [6-8]. Recently, with the growing dominance of deep learning (DL) in modern ML, various DL models have been modified and applied to vibration-based SHM [9-11]. In outlier analysis-based structural anomaly detection, deep generative model (DGM)-based approaches have gained popularity due to their unsupervised nature [12]. These methods utilize DGMs to model a complex, high-dimensional baseline distribution from training data to represent normal conditions. Compared to shallow learning-based methods, DGMs generally perform better at capturing the complex statistical properties of training data owing to the exceptional nonlinear mapping capability of deep neural networks [13]. Additionally, DGMs can be flexibly applied in supervised or semi-supervised learning given labeled training data, as they can extract representative features from raw structural response measurements [14]. Furthermore, leveraging the learned probability distributions, DGMs can be used for data augmentation to address the common challenge of limited training data in vibration-based SHM [15, 16]. Utilizing the physical properties of long-gauge static strain TF, Liu et al. [17] proposed an element-wise structural anomaly detection method using paralleled variational autoencoders (VAEs), which can also be used for damage quantification given labeled training data. Inspired by the representation learning capability of VAEs, Ma et al. [13] developed a VAE-based structural anomaly detection method, which combines a moving window with damage-sensitive features automatically extracted by a VAE to detect structural anomalies. Luo et al. [18] introduced an improved generative adversarial network (GAN) to extract latent representations from raw data and integrated it with cloud model theory to reduce damage misjudgment and enhance the robustness of structural anomaly detection.

Despite these achievements, outlier analysis-based methods are limited to a binary distinction between “healthy” and “damaged”. In practice, however, structural responses typically exhibit multiple patterns due to environmental and operational variabilities (EOVs) and structural damage. This mismatch makes outlier analysis-based methods sensitive to threshold establishment and unable to provide additional information about the operating conditions of the monitored structure [19, 20]. Clustering offers a promising solution to this limitation, which partitions the dataset into clusters in an unsupervised manner based on the inherent similarity of data points, with each cluster corresponding to a distinct pattern of structural behavior caused by different environmental and operational conditions or damage modes. By identifying these patterns, clustering provides deeper insights into structural behavior, thus enabling more accurate modeling of normal operating conditions to improve the robustness of anomaly detection [21] and guiding decision-makers in selecting appropriate follow-up actions for detected structural anomalies [19, 20]. Moreover, information identified by clustering can be integrated into semi-supervised learning for higher-level SHM tasks beyond anomaly detection, such as damage classification [22, 23]. Consequently, clustering-based methods have attracted increasing attention in vibration-based structural anomaly detection and health monitoring in recent years [9]. Representative approaches, such as K-means clustering [24], hierarchical clustering [25], and Gaussian mixture model (GMM) clustering [26], have demonstrated the feasibility and effectiveness of clustering in this field. However, most of these methods require the number of clusters to be prespecified [25], but there is no standard method to determine the optimal cluster number [27]. Additionally, most existing methods lack incremental learning capability, which refers to the ability to process new data in real-time without forgetting previously learned information [28] and is demanded for practical structural anomaly detection systems [19, 20]. To address these issues, Bayesian nonparametric mixture models have been introduced in structural anomaly detection and health monitoring [19, 20], which employ a nonparametric prior to dynamically adapt the cluster number based on observed data while enabling incremental learning by updating the posterior with new observations. However, these approaches rely on shallow learning, limiting their capability to handle high-dimensional, highly nonlinear raw structural responses. Therefore, they require a manual feature selection process for dimensionality reduction [19, 20], making their performance sensitive to the quality of these manually-engineered structural features.

In light of the aforementioned issues and challenges, this work proposes a deep clustering framework, referred to as Dirichlet process-deep generative model-integrated incremental learning (DPGIIL), for TF-based online structural anomaly detection. While this study uses a VAE-based implementation, referred to as Dirichlet process-variational autoencoder-integrated incremental learning (DPVIIL), as an illustrative example, this framework can be generalized to other DGMs, such as flow-based models [29]. Although DGM-based clustering has not been widely adopted in vibration-based structural anomaly detection, it has gained attention in recent years.

Some studies [30-32] integrated GMMs into DGMs to utilize the GMM as a prior in the latent space for enhancing information capacity, while the DGMs' representation learning capabilities overcome the limitations of traditional GMMs in capturing nonlinearity [33]. To address the need for prespecifying the cluster number, Nalisnick and Smyth [34] introduced the stick-breaking VAE by replacing the prior of the vanilla VAE with a stick-breaking process. While this method effectively learns discriminative latent representations for clustering, it lacks detailed information about individual cluster shapes and densities. Bing et al. [35] proposed a deep clustering method by integrating memoized online variational inference-based Dirichlet process mixture models (DPMMs) [36] into the VAE framework. Despite its good performance, this method lacks a generative process for latent representations and a mathematically rigorous analysis of the loss function. Different from existing methods, DPVIIL introduces a DPMM prior into the latent space of the VAE for clustering, enabling the cluster number to adapt dynamically based on the observed data. It also provides a mathematically solid framework to jointly optimize the VAE and the DPMM based on a novel lower bound on its log marginal likelihood, which allows the DPMM to regularize the VAE's feature extraction process and promotes the VAE to learn discriminative representations from raw data for enhanced clustering performance. Additionally, DPVIIL can be trained incrementally, allowing the model to be updated continuously as new data becomes available. For TF-based online structural anomaly detection, DPVIIL automatically extracts informative latent representations from raw TF vectors and clusters them to reveal underlying patterns of structural behavior. The structural condition can be determined based on the clustering results [19]: Initially, a set of TF vectors from normal operating conditions is used to train DPVIIL and identify baseline clusters representing healthy states. During the subsequent monitoring phase, as new TF vectors gradually arrive, DPVIIL is incrementally updated and dynamically creates new clusters to adapt to evolving data. Anomaly detection is performed by determining whether a new sample is assigned to one of the baseline clusters representing normal conditions or to a newly created cluster indicating an abnormal state. A schematic illustration of DPVIIL and its application in TF-based online structural anomaly detection is presented in Fig. 1, and the main contributions of this work are summarized as follows:

- A deep clustering framework is developed for online structural anomaly detection by integrating the DPMM and DGM, which leverages the representation learning capability of DGMs to avoid manual feature selection required in traditional DPMM-based clustering, combined with the DPMM's advantage in adaptively identifying distinct patterns in observed data.
- A lower bound on the log marginal likelihood of DPVIIL, tighter than the evidence lower bound, is derived to enable the joint optimization of both DPMM and network parameters through an iterative approach. This method allows the DPMM to regularize the feature extraction process of the VAE, while the latent representations concurrently influence the clusters identified by the DPMM. Consequently, DPVIIL enables dynamically adjusting the number, density, and shape of clusters in the latent space of the VAE based on observed data.

- By retaining the summary statistics of the DPMM and the network parameters, DPVIIL enables adapting the identified clusters within an incremental learning framework and dynamically generating new clusters to indicate previously unseen structural conditions, making it well-suited for online structural anomaly detection.

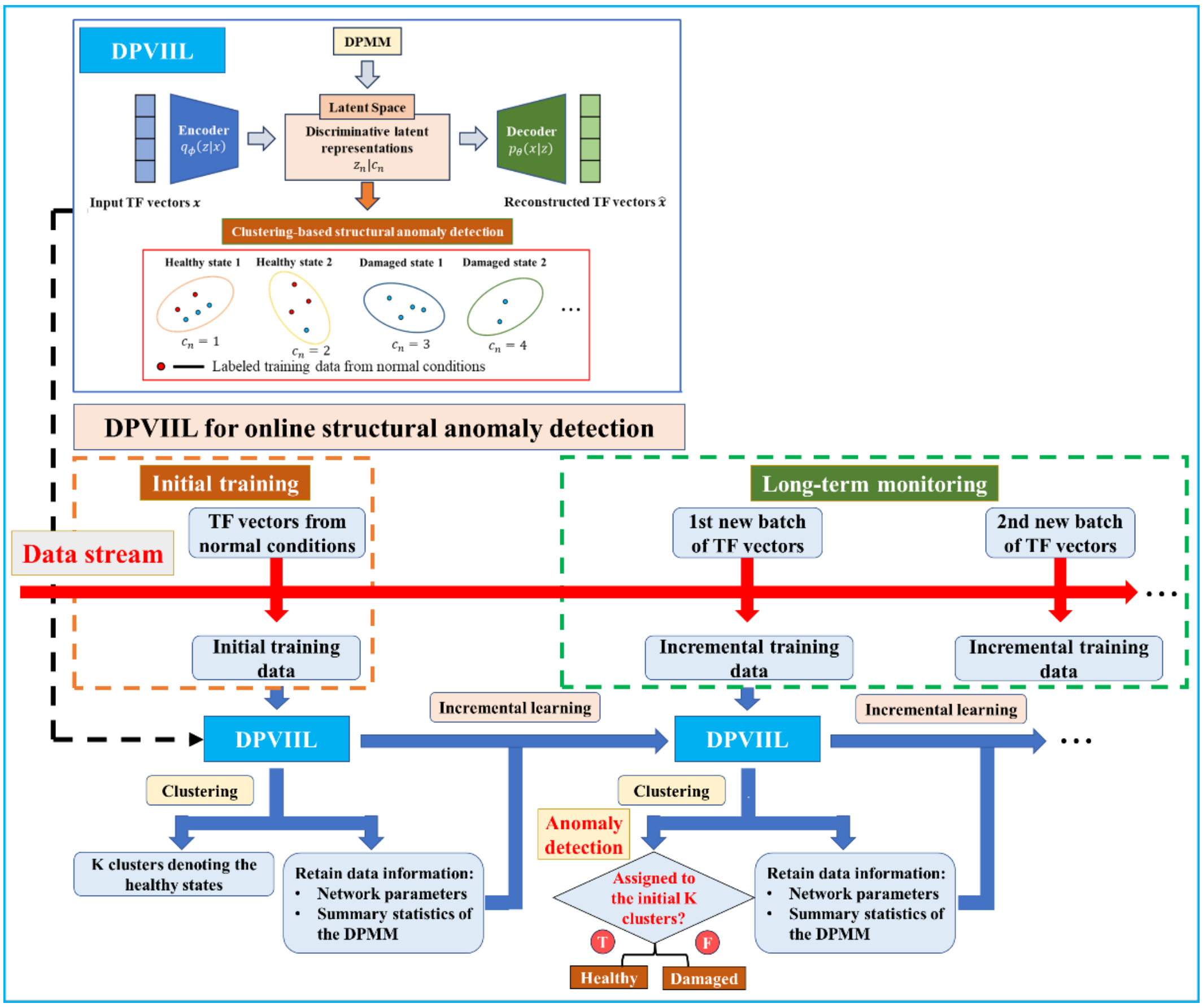

Figure 1. Schematic illustration of DPVIIL and its application to online structural anomaly detection.

## 2 Preliminaries

In this section, some key concepts of DPMMs and the VAE are provided, which form the theoretical foundation of the proposed deep clustering framework. Additionally, the rationale for applying this framework to TF-based online structural anomaly detection is illustrated.

## 2.1 Dirichlet process mixture models

### *2.1.1 Fundamentals of the DPMM*

The DPMM is a Bayesian nonparametric mixture model used for density estimation and clustering, which employs the Dirichlet process (DP) as a nonparametric prior to encompass both the component number and the parameters of each component. The DP is a stochastic process parameterized by a concentration parameter $\alpha$ and a base distribution $H$. Each sample of a DP is a discrete distribution whose marginal distributions are Dirichlet distributed [37]. For the DPMM, the DP is used as a nonparametric prior in a hierarchical Bayesian specification, with the explicit realization achieved through the stick-breaking representation [38]: In each step, a random variable $v_k$ is drawn from the beta distribution $\mathrm{Beta}(1,\alpha)$, and the mixing proportion of each component is $\pi_k = \pi_k(v) = v_k \prod_{i=1}^{k-1}(1-v_i)$. This can be metaphorically interpreted as successively breaking a unit length stick into infinite segments, as shown in Fig. 2 [37]. Subsequently, the parameters of each component $\eta_k$ are drawn from the base distribution $H$, while the generative process of an observation $x_n$ is determined by these parameters and an assignment variable $c_n$ drawn from the categorical distribution $\mathrm{Cat}(\pi_k(v))$, i.e., $x_n \sim F(x_n|\eta_{c_n})$ and $c_n \sim \mathrm{Cat}(\pi_k(v))$. In DPMM-based clustering, each cluster corresponds to a mixture component in the DPMM, and the number of clusters is inferred from the observed data rather than fixed in advance. This attribute makes it well-suited for structural anomaly detection and health monitoring, where structural responses typically exhibit multiple patterns due to EOVs and different damage scenarios, while the true number of such patterns is difficult to determine [19, 20]. The distribution of data within the $k$-th cluster is denoted by $F(x|\eta_k)$, where $\eta_k$ is drawn from the base distribution $H$. The probability of assigning the data point $x_n$ to the $k$-th cluster, also known as the responsibility, can be computed similarly to traditional GMMs, i.e., $p(c_n = k) = \pi_k F(x_n|\eta_k) \Big/ \sum_{j=1}^{\infty} \pi_j F(x_n|\eta_j)$. Conditioned on a set of observed data $X$, the posterior responsibilities $p(c_n = k|X)$ can be estimated to assign data points to clusters, thereby enabling the DPMM to perform flexible clustering. For more details of the DPMM, one can refer to [37, 38].

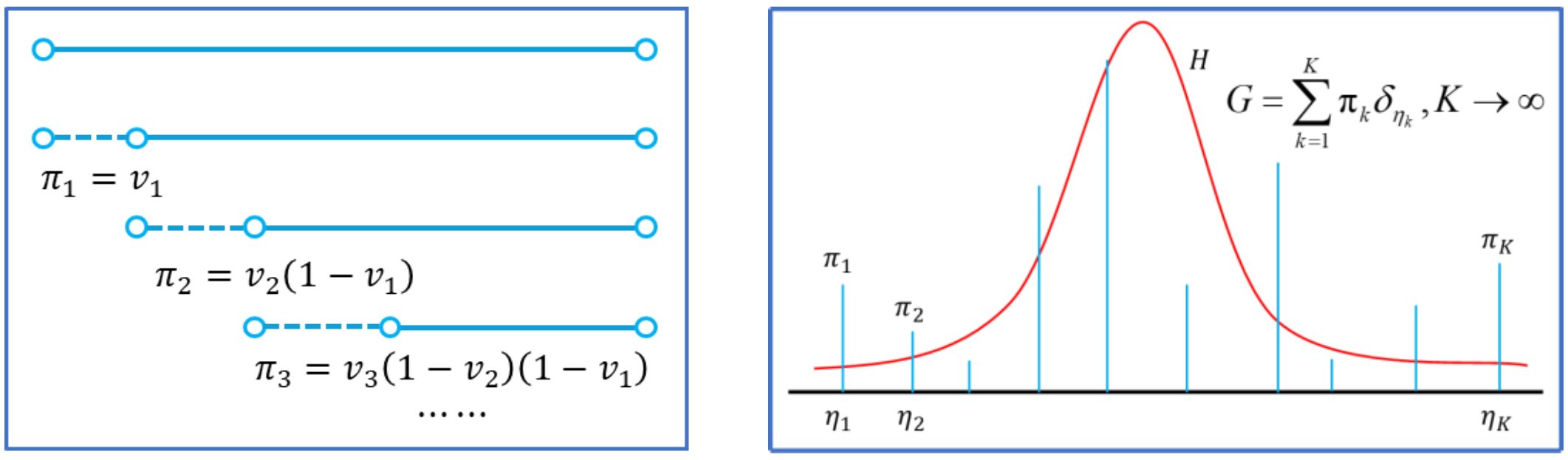

(a) (b)

Figure 2. Schematic illustration of (a) the stick-breaking representation and (b) a sample drawn from a Dirichlet process.

### *2.1.2 Coordinate ascent variational inference (CAVI) for the DPMM*

Like many other Bayesian approaches, explicit inference for the true posterior distribution under a DPMM prior is challenging, which motivates the development of CAVI-based methods [19, 36, 38, 39] to approximate the true posterior. These methods use the conjugate prior of component parameters as the base distribution, combined with a fully factorized variational distribution, to iteratively optimize each parameter. In this work, the truncation-free VI-DPGMM [19] is followed as it explicitly accounts for inactive components (those without assigned data) in the DPMM to provide a more reasonable criterion for adding new active components to the variational distribution. Assuming the base distribution is $H(\eta|\varphi)$, the posterior of the DPMM can be derived using Bayes' rule:

$$\begin{aligned} p(v,c,\eta|X,\varpi) &= \frac{p(v,c,\eta|\varpi)\,p(X|v,c,\eta)}{p(X|\varpi)} \\ &= \frac{\prod_{k=1}^{\infty}\mathrm{Beta}(v_k\,|\,1,\alpha)\,H(\eta_k|\varphi)\prod_{n=1}^{N}\mathrm{Cat}(c_n\,|\,v)\,F(x_n\,|\,\eta_{c_n})}{p(X|\varpi)} \end{aligned} \tag{1}$$

where $X=\{x_1,x_2,...,x_N\}$ is observed data; $\varpi=\{\alpha,\varphi\}$ denotes the hyperparameters; both $F(\bullet)$ and $H(\bullet)$ are assumed to belong to the exponential family. In terms of VI, a variational distribution $q(v,c,\eta|\hat{\varpi})$ is introduced to approximate the true posterior by minimizing the Kullback-Leibler (KL) divergence between them, which is equivalent to maximizing the evidence lower bound (ELBO) $\mathcal{L}(\hat{\varpi})$ given as [19]:

$$\mathcal{L}(\hat{\varpi})=\mathbb{E}_q\left[\log p(X,v,c,\eta|\varpi)\right]-\mathbb{E}_q\left[\log q(v,c,\eta|\hat{\varpi})\right] \tag{2}$$

Based on the mean-field assumption, a fully factorized variational distribution $q(v,c,\eta|\hat{\varpi})=\prod_{k=1}^{\infty}q(v_k|\hat{\alpha}_k)\prod_{k=1}^{\infty}q(\eta_k|\hat{\varphi}_k)\prod_{n=1}^{N}q(c_n|\hat{\pi}_n)$ is used, where $\hat{\varpi}=\{\hat{\alpha},\hat{\varphi},\hat{\pi}\}$ denotes the variational parameters and $q(v_k|\hat{\alpha}_k)=\mathrm{Beta}(v_k|\hat{\alpha}_{k1},\hat{\alpha}_{k2})$, $q(\eta_k|\hat{\varphi}_k)=H(\eta_k|\hat{\varphi}_k)$, $q(c_n|\hat{\pi}_n)=\mathrm{Cat}(c_n|\hat{\pi}_{n1},\hat{\pi}_{n2},...,\hat{\pi}_{nK})$. To handle the infinite components in the DPMM, [19] introduced a truncation-free VI by setting the variational distributions of the parameters for inactive components as their corresponding priors. Based on this assumption, the sum of the probabilities for a data point to be assigned to these inactive components can be analytically derived:

$$\sum_{k=K_a+1}^{\infty} \hat{\pi}_{nk} = \frac{\sum_{k=K_a+1}^{\infty} \hat{\rho}_{nk}}{\sum_{j=1}^{K_a} \hat{\rho}_{nj} + \sum_{j=K_a+1}^{\infty} \hat{\rho}_{nj}}; \sum_{k=K_a+1}^{\infty} \hat{\rho}_{nk} = \frac{\hat{\rho}_{n(K_a+1)}}{1-\exp\left\{\psi(\alpha)-\psi(1+\alpha)\right\}} \tag{3}$$

where $K_a$ is the number of active components; $\hat{\rho}_{nk}$ is the unnormalized form of $\hat{\pi}_{nk}$; $\psi(\bullet)$ denotes the digamma function. Given the updated $\hat{\pi}_{nk}$, the other variational parameters, i.e., $\hat{\alpha}_k$ and $\hat{\varphi}_k$, can also be optimized using the coordinate ascent algorithm [38], and the ELBO is given by:

$$\mathcal{L}(\hat{\varpi}) = \sum_{k=1}^{K_a}\left(\mathbb{E}_q\left[\log\frac{p(v_k \mid \alpha)}{q(v_k \mid \hat{\alpha}_k)}\right] + \mathbb{E}_q\left[\log\frac{p(\eta_k \mid \varphi)}{q(\eta_k \mid \hat{\varphi}_k)}\right]\right) + \sum_{n}^{N}\log\sum_{k=1}^{\infty}\hat{\rho}_{nk} \tag{4}$$

The sum of probabilities $\sum_{k=K_a+1}^{\infty} \hat{\pi}_{nk}$ then serves as the criterion to activate new components [19], allowing this method to dynamically adjust the component number to adapt to observed data. For more details about this algorithm, one can refer to [19].

Despite the advantages of the truncation-free VI-DPGMM, its shallow learning nature limits its ability to handle high-dimensional, complex data such as TF vectors. As a result, its performance is sensitive to the quality of manually engineered low-dimensional features when employed in structural anomaly detection. To address this issue, this work proposes a deep clustering framework by integrating the DPGMM into DGMs, which can simultaneously extract informative features from raw dynamic responses and cluster them to indicate different structural conditions. This framework is implemented using a VAE as an illustrative example, while it can also be generalized to other DGMs.

### 2.2 Variational autoencoders

The VAE [40], as illustrated in Fig. 3, is one of the most prevalent DGMs, which connects a neural network (encoder) to another one (decoder) through a probabilistic latent space. The latent space is typically modeled as a diagonal Gaussian distribution that corresponds to the parameters of a variational distribution. Despite its strong performance in probabilistic modeling and feature extraction, the vanilla Gaussian VAE struggles to learn discriminative representations from raw data, as the latent representation for each sample is modeled with a distinct distribution, limiting its ability to capture shared structures within the dataset [30]. This limitation restricts the primary application of vanilla VAEs in structural anomaly detection to outlier analysis-based methods, which are sensitive to threshold establishment and fail to provide additional information about the operating conditions of the monitored structure. To address this issue, this work introduces a DPMM prior into the latent space of vanilla VAEs to form DPVIIL, which enables extracting discriminative representations from raw dynamic responses for clustering-based structural anomaly detection.

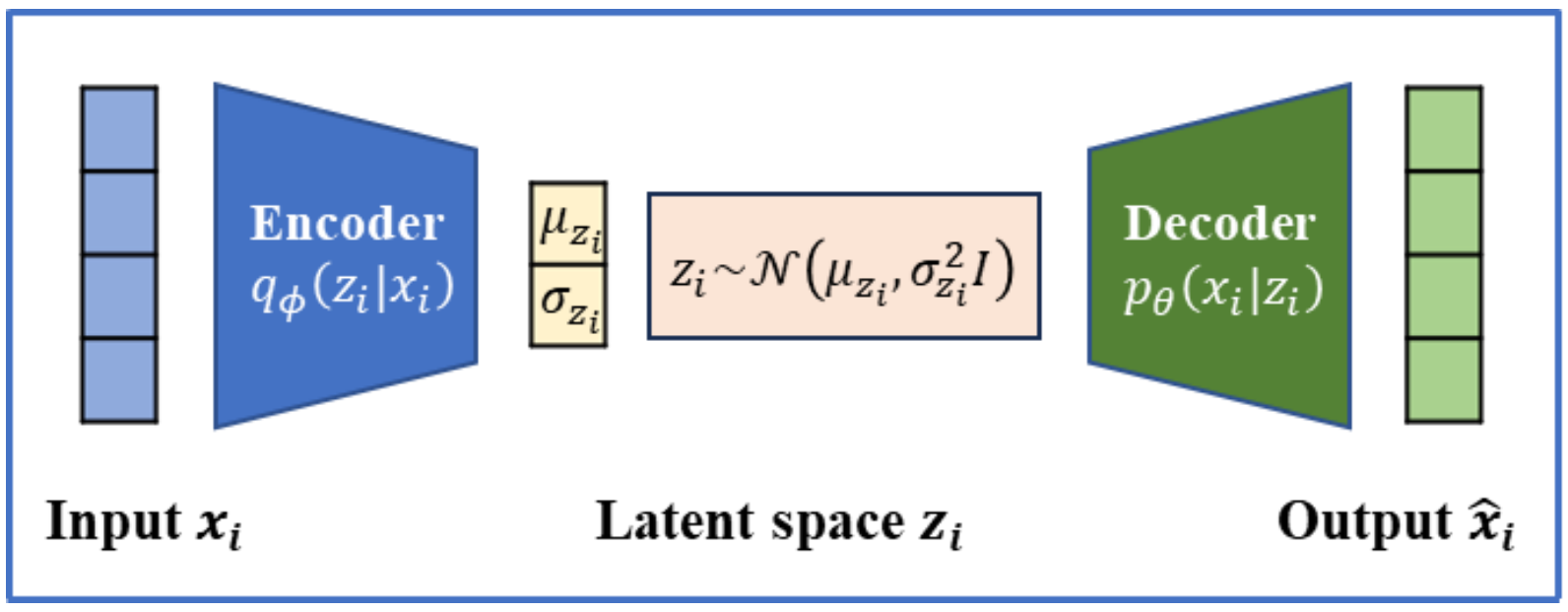

Figure 3. Schematic diagram of a vanilla VAE.

### 2.3 Rationale of DPVIIL in TF-based online structural anomaly detection

By integrating the DPMM with the VAE, DPVIIL automatically extracts discriminative representations from raw data and partitions them into clusters for online structural anomaly detection. In this work, TF vectors are used as inputs to leverage the sensitivity of TFs to structural damage and their robustness to excitation [41-43]. Each cluster identified by DPVIIL groups TF vectors with similar statistical properties. Differences between clusters capture systematic changes in TFs, such as shifts in their amplitudes or peak frequencies, that alter the statistical properties of TF vectors and thus reflect changes in the frequency domain characteristics of the monitored structure. To distinguish between clusters representing normal operating conditions and those indicating the existence of structural damage, a clustering-based decision criterion is adopted in this work: In the initial training phase, a set of TF vectors from healthy states (available a priori) is used to train DPVIIL, yielding several clusters that represent normal operating conditions. During subsequent long-term monitoring, DPVIIL is incrementally updated as new batches of TF vectors arrive, while clustering proceeds without interruption. If a new data point is assigned to one of the clusters generated in the initial training phase, it corresponds to a normal operating condition; if it is assigned to a newly created cluster, it indicates a structural anomaly potentially caused by damage. Compared to existing approaches, DPVIIL performs end-to-end representation learning and clustering in a unified framework and avoids sensitivity to manual feature engineering. It also forms clusters to indicate different structural conditions, providing more information about the operating conditions of the monitored structure than outlier analysis-based methods that merely distinguish between "healthy" and "damaged" states. Furthermore, DPVIIL can be dynamically updated through incremental learning, making it well-suited for online structural anomaly detection during long-term monitoring.

## 3 Dirichlet process-variational autoencoder-integrated incremental learning

This section provides a detailed description of DPVIIL, a probabilistic clustering model that combines DPMM and VAE. In this work, DPVIIL extracts latent representations from raw TF vectors through a probabilistic encoder and clusters them by introducing a DPMM into the latent space to identify distinct structural patterns. Simultaneously, a probabilistic decoder is used to reconstruct the TF vectors based on latent representations drawn from the mixture distribution modeled by the DPMM. Similar to vanilla VAEs [40], the decoder, conditioned on a DPMM prior, models the generative process of raw TF vectors measured from different structural conditions influenced by EOVs or damage, while the encoder, along with the inferred posterior of the DPMM, approximates the true posterior of the generative model through variational inference. This structure enables the DPMM to regularize the latent space of the VAE, promoting DPVIIL to extract discriminative representations from raw TF vectors for clustering. A schematic illustration of DPVIIL and its application to TF-based structural anomaly detection is presented in Fig. 4, with the symbols defined in Section 2.

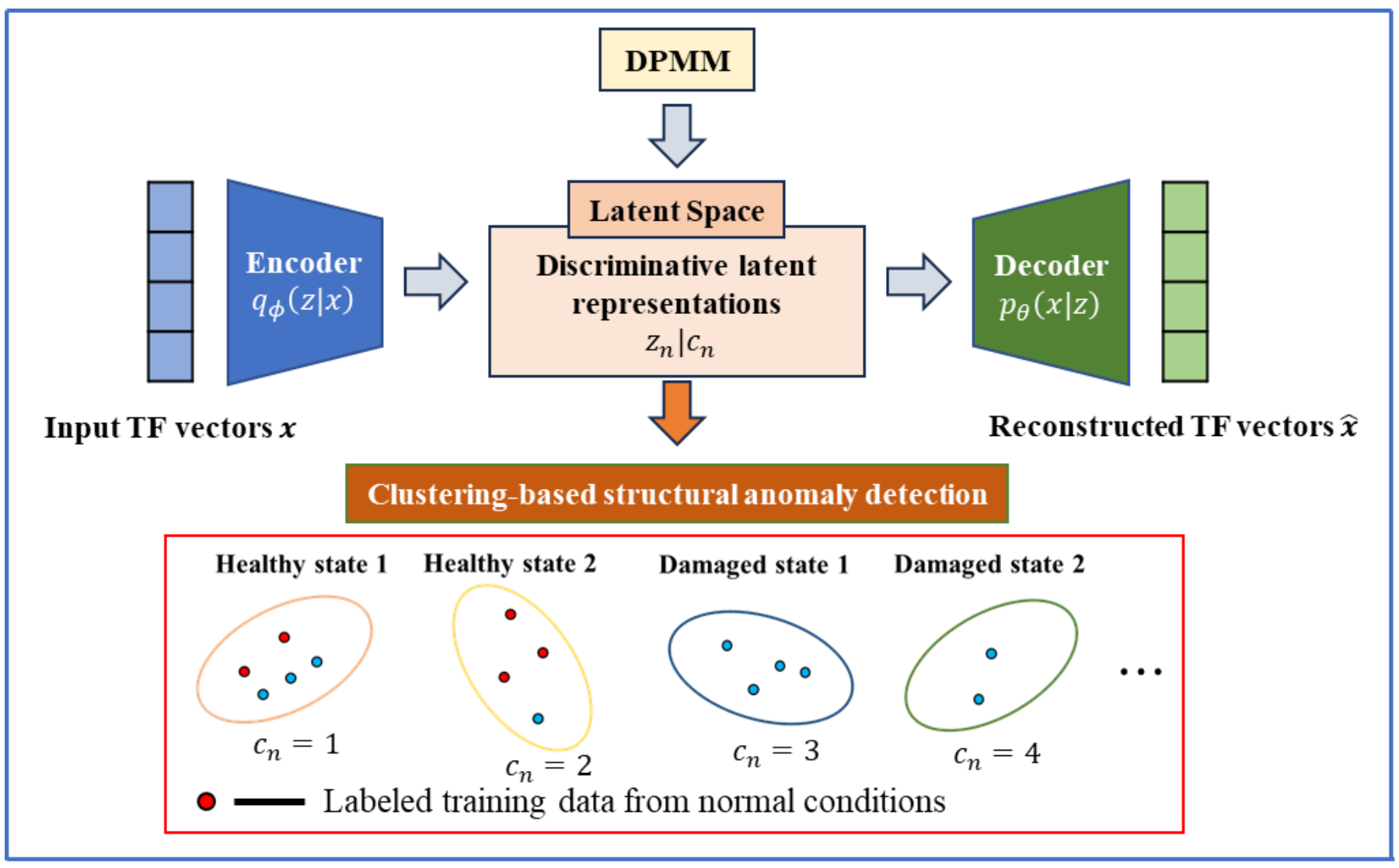

Figure 4. Schematic diagram of DPVIIL and its application in structural anomaly detection.

### 3.1 The generative process

Since DPVIIL is an unsupervised generative approach to clustering, the random process through which raw TF vectors are generated is introduced first. Specifically, an observed sample $x_n \in \Re^D$ is generated by the following process within the context of DPVIIL, with a schematic diagram provided in Fig. 5 to facilitate illustration.

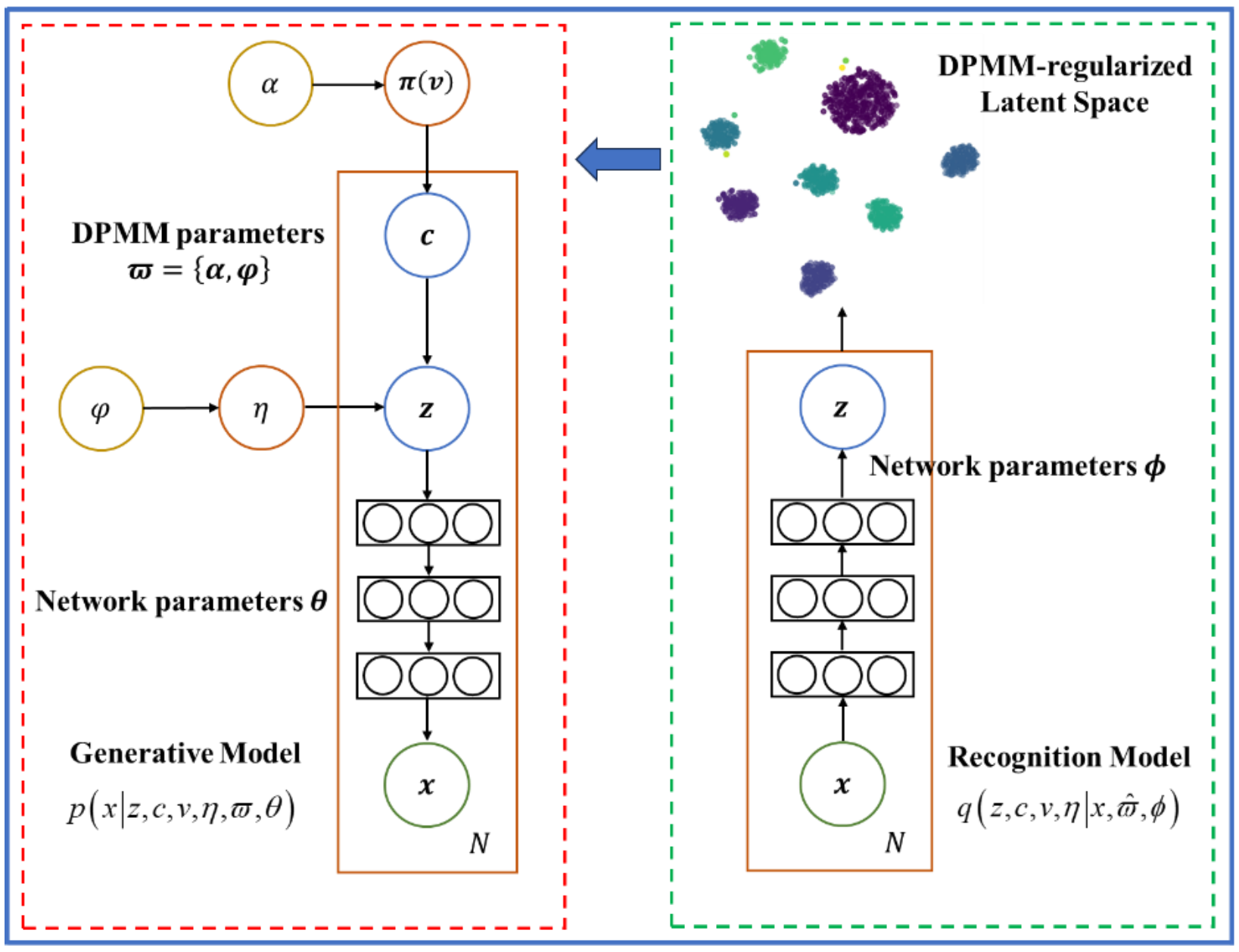

Figure 5. Schematic illustration of the generative process of DPVIIL.

| Generative process of DPVIIL: |
| --- |

1. Draw a sample $v_k \sim \mathrm{Beta}(1,\alpha)$
2. Draw an assignment variable $c_n \sim \mathrm{Cat}(\pi_k(v))$, $\pi_k(v) = v_k \prod_{i=1}^{k-1}(1-v_i)$, where $\mathrm{Cat}(\bullet)$ is the categorical distribution
3. Draw a sample of component parameters $\eta \sim H(\eta|\varphi)$
4. Draw a latent variable $z_n|c_n \sim F(z|\eta_{c_n})$

5. Draw a sample $x_n$ (here we assume $x_n$ is real-valued):

   1) Compute the mean and standard deviation $[\mu_x, \sigma_x] = f(z, \theta)$, where $f(\bullet)$ is a function modeled by a neural network

   2) Draw a sample $x_n \sim \mathcal{N}(\mu_x, \sigma_x^2 I)$

Based on this generative model, the objective is to estimate the posterior distribution of its parameters $p(z, c, v, \eta | x, \varpi, \theta)$ to mimic the underlying random process through which the TF vectors are generated. Specifically, the true posterior can be expressed as:

$$
\begin{aligned}
p(z, c, v, \eta, \theta | \varpi, x) &= \frac{p(z, c, v, \eta | \varpi) p(x | z, \theta)}{\log p(x | \varpi, \theta)} \\
&= \frac{p(v | \alpha) p(c | v) p(\eta | \varphi) p(z | c, \eta) p(x | z, \theta)}{\log p(x | \varpi, \theta)}
\end{aligned} \tag{5}
$$

Here $F(\bullet)$ is assumed to be Gaussian and $H(\bullet)$ is set as the corresponding conjugate prior, then it has:

$$
\begin{aligned}
& p(v | \alpha) = \text{Beta}(1, \alpha);\ p(c | v) = \text{Cat}(\pi_k(v)) \\
& H(\eta | \varphi) = p(\mu, \mathbf{\Lambda} | \varphi) = \mathcal{NW}(\mu, \mathbf{\Lambda} | m, \lambda, \text{W}, \upsilon) \\
& F(z | \eta_c) = p(z | c, \eta) = \mathcal{N}\left(z | \mu_c, (\mathbf{\Lambda}_c)^{-1}\right) \\
& p(x | z, \theta) = \mathcal{N}(\mu_x, \sigma_x^2 I);\ [\mu_x, \sigma_x] = f(z, \theta)
\end{aligned} \tag{6}
$$

where $\mathcal{NW}(\bullet)$ is the normal-Wishart distribution; $\mu$ and $\mathbf{\Lambda}$ denote the mean vector and precision matrix of the Gaussian distribution, respectively.

## 3.2 Variational inference for DPVIIL

### *3.2.1 Tighter lower bound on the log marginal likelihood of DPVIIL*

As the true posterior is analytically intractable, a variational distribution, also referred to as the recognition model [40], $q(z, c, v, \eta | x, \hat{\varpi}, \phi)$ is introduced to approximate the true posterior by minimizing the KL divergence between them, which

is given by:

$$
\begin{aligned}
&D_{KL}\left(q\left(z,c,v,\eta|x,\hat{\varpi},\phi\right)\|p\left(z,c,v,\eta,\theta|\varpi,x\right)\right) \\
&=\log p\left(x|\varpi,\theta\right)-\mathbb{E}_{q\left(z,c,v,\eta|x,\hat{\varpi},\phi\right)}\left[\log\frac{p\left(x,z,c,v,\eta,\theta|\varpi\right)}{q\left(z,c,v,\eta|x,\hat{\varpi},\phi\right)}\right] \\
&\triangleq \log p\left(x|\varpi,\theta\right)-\mathcal{L}_{ELBO}\left(\theta,\phi,\hat{\varpi}|x\right)
\end{aligned}
\tag{7}
$$

Therefore, minimizing the KL divergence is equivalent to maximizing the ELBO $\mathcal{L}_{ELBO}\left(\theta,\phi,\hat{\varpi}|x\right)$ by optimizing the parameters $\theta,\phi,\hat{\varpi}$, which can be rewritten as:

$$
\mathcal{L}_{ELBO}\left(\theta,\phi,\hat{\varpi}|x\right)=\mathbb{E}_{q(z|x,\phi)}\left[\log p\left(x|z,\theta\right)\right]-D_{KL}\left[q\left(z,c,v,\eta|x,\hat{\varpi},\phi\right)\|p\left(z,c,v,\eta|\varpi\right)\right] \tag{8}
$$

where $\hat{\varpi}$ denotes the variational parameters of the DPMM and is defined in Section 2. To facilitate optimization, a fully factorized variational distribution is adopted based on the mean-field assumption, and the variational distribution for the DPMM is chosen to belong to the same family as the true posterior distribution [19]. As a result, the overall variational distribution can be expressed as:

$$
q\left(z,c,v,\eta|x,\hat{\varpi},\phi\right)=q\left(z|x,\phi\right)q\left(c,v,\eta|z,\hat{\varpi}\right)=q\left(z|x,\phi\right)q\left(v\,|\,z,\hat{\alpha}\right)q\left(c\,|\,z,\hat{\pi}\right)q\left(\eta|z,\hat{\varphi}\right) \tag{9}
$$

where each factor can be expressed as $q\left(v\,|\,z,\hat{\alpha}\right)=\mathrm{Beta}\left(\hat{\alpha}_1,\hat{\alpha}_2\right)$, $q\left(c\,|\,z,\hat{\pi}\right)=\mathrm{Cat}\left(\hat{\pi}_k\right)$, and $q\left(\eta|z,\hat{\varphi}\right)=q\left(\mu,\mathbf{\Lambda}|z,\hat{\varphi}\right)=\mathcal{NW}\left(\mu,\mathbf{\Lambda}|\hat{m},\hat{\lambda},\hat{\mathrm{W}},\hat{\upsilon}\right)$. Additionally, similar to vanilla VAEs [40], $q\left(z|x,\phi\right)$ is modeled using a Gaussian neural network $g$:

$$
\left[\mu_z,\sigma_z\right]=g\left(x,\phi\right);\ z\sim\mathcal{N}\left(\mu_z,\sigma_z^2 I\right) \tag{10}
$$

For a single data point $x_n$, the first term in Eq. (8), commonly referred to as the reconstruction loss, can be estimated using the stochastic gradient variational Bayes (SGVB) estimator and the reparameterization trick [40]:

$$
\mathbb{E}_{q\left(z_n|x_n,\phi\right)}\left[\log p\left(x_n|z_n,\theta\right)\right]\approx\frac{1}{L}\sum_{l=1}^{L}\log p\left(x_n\left|z_n^{(l)},\theta\right.\right) \tag{11}
$$

where $z_n^{(l)}$ is a sample drawn from the variational distribution, i.e., $z_n^{(l)} \sim q\left(z_n | x_n, \phi\right)$, and is given by [40]:

$$z_n^{(l)} = g_\phi\left(\epsilon_n^{(l)}, x_n\right) = \mu_{z_n} + \sigma_{z_n} \odot \epsilon_n^{(l)};\ \epsilon^{(l)} \sim \mathcal{N}\left(0, I\right) \tag{12}$$

Based on Eqs. (11)-(12), the reconstruction loss can be estimated. However, the KL divergence term $D_{KL}\left[q\left(z,c,v,\eta | x,\hat{\varpi},\phi\right) \| p\left(z,c,v,\eta | \varpi\right)\right]$ is analytically intractable as it involves a complex partition of the latent space through a DPMM. To address this, a tractable lower bound on the log marginal likelihood of DPVIIL is derived to serve as the objective function for jointly optimizing the DPMM and network parameters. This bound is tighter than the ELBO given sufficient training data and an appropriate penalizing hyperparameter $\gamma$, and can be expressed as follows:

$$\begin{aligned} \mathcal{L}\left(\theta,\phi,\hat{\varpi} | x\right) = \mathbb{E}_{q(z|x,\phi)}\left[\log p\left(x | z,\theta\right)\right] + \mathbb{E}_{q(c,v,\eta|\hat{\varpi})}\left[\log \frac{p\left(c,v,\eta \mid \varpi\right)}{q\left(c,v,\eta \mid z,\hat{\varpi}\right)}\right] \\ -\gamma \sum_{k=1}^{\infty} q\left(c=k\right) D_{KL}\left[q\left(z | x,\phi\right) \| \mathcal{N}\left(z \middle| \hat{m}_k, \left(\hat{\lambda}_k \hat{\mathrm{W}}_k\right)^{-1}\right)\right] \end{aligned} \tag{13}$$

The detailed derivation of this tighter lower bound is provided in Appendix I.

#### *3.2.2 Two-step iterative optimization approach for inferring DPVIIL*

Although the lower bound given in Eq. (13) is tractable, optimizing $\theta$, $\phi$, and $\hat{\varpi}$ simultaneously remains challenging, as the DPMM involves a countably infinite number of components [37]. To address this issue, a two-step iterative approach is proposed to jointly optimize the variational parameters of the DPMM, i.e., $\hat{\varpi}$, and the parameters of the neural networks, i.e., $\theta$ and $\phi$, given the training set $X = \{x_n\}_{n=1}^{N}$. In each iteration, this approach firstly updates $\hat{\varpi}$ through CAVI using extracted latent representations, with the number of active components inferred automatically. Then, $\theta$ and $\phi$ are updated using the SGVB method and the reparameterization trick, while keeping the DPMM parameters fixed and omitting the inactive components. This iterative process can be summarized as follows:

(1) Fix $\theta$ and $\phi$, and optimize $\hat{\varpi}$ using a CAVI optimizer with a greedy split-merge scheme based on the extracted latent representations $Z=\{z_n\}_{n=1}^{N}$, as detailed in the next section. In this step, the objective becomes maximizing the ELBO with respect to $\hat{\varpi}$, which is given by $\mathcal{L}_{CAVI}\left(\hat{\varpi}|Z\right)=\mathbb{E}_{q(c,v,\eta|\hat{\varpi})}\left[\log\frac{p\left(Z,c,v,\eta\mid\varpi\right)}{q\left(c,v,\eta\mid Z,\hat{\varpi}\right)}\right]$. This step automatically infers the number of active components $K_a$ and yields the optimized variational parameters of the DPMM, i.e., $\hat{\varpi}=\{\hat{\alpha}_k,\hat{\varphi}_k,\hat{\pi}_{nk}\}_{k=1,2,\ldots,K_a}^{n=1,2,\ldots,N}$, which allow for calculating the lower bound given in Eq. (13) by omitting the inactive components when optimizing the network parameters in the next step.

(2) Fix the DPMM parameters $\hat{\varpi}$ and the number of active components $K_a$, and optimize $\theta$ and $\phi$ using the SGVB estimator and the reparameterization trick [40]. In this scenario, the term $\mathbb{E}_{q(c,v,\eta|z,\hat{\varpi})}\left[\log\frac{q\left(c,v,\eta\mid z,\hat{\varpi}\right)}{p\left(c,v,\eta\mid\varpi\right)}\right]$ in Eq. (13) can be ignored as it is irrelevant to $\theta$ and $\phi$, and the objective function becomes:

$$\begin{aligned}\mathcal{L}\left(\theta,\phi,\hat{\varpi}|X\right)&=\mathcal{L}\left(\theta,\phi|\hat{\varpi},X\right)=\sum_{n=1}^{N}\mathcal{L}\left(\theta,\phi|\hat{\varpi},x_n\right)\\&=\sum_{n=1}^{N}\mathbb{E}_{q(z_n|x_n,\phi)}\left[\log p\left(x_n|z_n,\theta\right)\right]-\gamma\sum_{k=1}^{K_a}q(c=k)D_{KL}\left[q\left(z_n|x_n,\phi\right)\middle\|\mathcal{N}\left(z_n\middle|\hat{m}_k,\left(\hat{\lambda}_k\hat{\mathrm{W}}_k\right)^{-1}\right)\right]\end{aligned}\tag{14}$$

where $\mathbb{E}_{q(z_n|x_n,\phi)}\left[\log p\left(x_n|z_n,\theta\right)\right]$ is given in Eq. (11). Given a minibatch of data $X_B=\{x_n\}_{n=1}^{B}$ randomly drawn from the training set, the objective function can be approximated by $\mathcal{L}\left(\theta,\phi|\hat{\varpi},X\right)\simeq\mathcal{L}^B\left(\theta,\phi|\hat{\varpi},X^B\right)=\frac{N}{B}\sum_{n=1}^{B}\mathcal{L}\left(\theta,\phi|\hat{\varpi},x_n\right)$ [40].

By alternating these two steps in each training epoch, the DPMM parameters $\hat{\varpi}$, the number of active components $K_a$, and the network parameters $\theta$ and $\phi$, can be iteratively updated until convergence. This joint optimization strategy allows the DPMM to regularize the feature extraction process of DPVIIL, thereby promoting the model to learn discriminative features from raw structural responses and to capture differences in their statistical properties to identify distinct structural conditions through clustering.

### 3.3 Greedy split-merge scheme-based CAVI for the DPMM

Given the latent representations $Z=\{z_n\}_{n=1}^{N}$ extracted from the training set, a modified CAVI optimizer is developed based on the method in [19] to estimate $K_a$ and $\hat{\varpi}$, which leverages the summary statistics of the DPMM [36] to retain information from previous data and enable incremental learning. Specifically, as the components of the DPMM are Gaussian distributions in this work, the summary statistics $s(Z)$ can be expressed as follows when omitting the inactive components:

$$s(Z)=\{s_k(Z)\}_{k=1}^{K_a}=\{N_k,\bar{z}_k,\mathrm{S}_k\}_{k=1}^{K_a} \tag{15}$$

where $N_k=\sum_{n=1}^{N}\hat{\pi}_{nk}$, $\bar{z}_k=\frac{1}{N_k}\sum_{n=1}^{N}\hat{\pi}_{nk}z_n$, $\mathrm{S}_k=\frac{1}{N_k}\sum_{n=1}^{N}\hat{\pi}_{nk}(z_n-\bar{z}_k)(z_n-\bar{z}_k)^T$. Conditioned on these summary statistics, the variational parameters can be updated given the prior, with detailed update equations provided in Appendix II. Then, a greedy split-merge scheme is proposed to determine $K_a$, which is summarized as follows:

(1) **Greedy split**: This step starts from $K_a=1$ and iteratively splits the component that maximally increases the ELBO $\mathcal{L}_{CAVI}(\hat{\varpi}|Z)$. Specifically, in each iteration, a component $k$ is split into two components $k_1$ and $k_2$ along the bisector of its principal component [39]. Subsequently, the summary statistics for these two components, denoted by $s_{k_1}(Z)$ and $s_{k_2}(Z)$, are computed accordingly, while the summary statistics for the other components remain fixed. The corresponding ELBO is then calculated and stored. This process is repeated for all active components to identify the split that yields the maximum ELBO. Afterward, the selected component is split, and the summary statistics for all components are updated to compute a new ELBO after the split, denoted by $\mathcal{L}_{CAVI}^{S}(\hat{\varpi}|Z)$. If the new ELBO exceeds the ELBO before splitting, i.e., $\mathcal{L}_{CAVI}^{S}(\hat{\varpi}|Z) > \mathcal{L}_{CAVI}(\hat{\varpi}|Z)$, the split is accepted with $K_a$ updated to $K_a = K_a + 1$, and the next iteration begins. Otherwise, the split is rejected, and the greedy split step is terminated. While this greedy strategy accelerates convergence compared to [19], it can be affected by outliers and may generate isolated components, leading to increased computational costs as training progresses [36]. To address this issue, a greedy merge step is introduced to mitigate local optima and reduce redundant isolated components.

(2) **Greedy merge**: This step iteratively merges two active components, $k_1$ and $k_2$, to maximize the ELBO using a greedy search. As randomly selecting $k_1$ and $k_2$ is computationally expensive and often ineffective, the method proposed in [36] is employed as a pre-selection strategy. Specifically, after selecting $k_1$ at random, $k_2$ is chosen to maximize the ratio of marginal likelihoods between the merged and separate configurations. Assume the merged component is denoted by $k_{12}$, its summary statistics is $s_{k_{12}}(Z) = s_{k_1}(Z) + s_{k_2}(Z)$, and the ratio of marginal likelihoods is given by [36]:

$$p(k_2|k_1) \propto \frac{M(s_{k_{12}}(Z))}{M(s_{k_1}(Z))M(s_{k_2}(Z))} \tag{16}$$

This ratio can be computed using the stored summary statistics, as described in [36], while the ELBO after merging can be derived based on $s_{k_{12}}(Z)$. In each iteration, the pair of components whose merge yields the maximum ELBO is selected and merged, and the corresponding ELBO, denoted by $\mathcal{L}_{CAVI}^{M}(\hat{\varpi}|Z)$, is computed and compared with the ELBO before merging. If $\mathcal{L}_{CAVI}^{M}(\hat{\varpi}|Z) > \mathcal{L}_{CAVI}(\hat{\varpi}|Z)$, the merge is accepted, and the next iteration is proceeds. Otherwise, the merge is rejected, and the greedy merge step is stopped.

Based on these two steps, the number of active components $K_a$ and the optimized variational parameters of the DPMM $\hat{\varpi} = \{\hat{\alpha}_k, \hat{\varphi}_k, \hat{\pi}_{nk}\}_{k=1,2,\ldots,K_a}^{n=1,2,\ldots,N}$ can be derived through a greedy split-merge scheme given the latent representations $Z$. To provide an intuitive illustration, the optimization process is summarized in Algorithm 1:

---

**Algorithm 1:** Greedy split-merge scheme-based CAVI for the DPMM

---

Require: latent representations $Z = \{z_n\}_{n=1}^{N}$.

Initialize: $K_a = 1$; prior parameters $\varpi = \{\alpha, \varphi\}$; initial summary statistics $s_k(Z)$ and lower bound $\mathcal{L}_{CAVI}(\hat{\varpi}|Z)$.

**A. Greedy split step:**

**Repeat:**

(A1) For each component $k \le K_a$, do:

- Split component $k$ into two components $k_1$ and $k_2$.
- Update the summary statistics $s_{k_1}(Z)$ and $s_{k_2}(Z)$, with the summary statistics for the other components fixed.
- Calculate the new ELBO $\mathcal{L}_{CAVI}^{S_k}(\hat{\varpi}|Z)$ and revert to the configuration before splitting.

(A2) Identify the component $K$ with $\mathcal{L}_{CAVI}^{S_K}(\hat{\varpi}|Z) = \max\{\mathcal{L}_{CAVI}^{S_k}(\hat{\varpi}|Z)\}_{k=1}^{K_a}$ and split it into two components.

(A3) Update $s_k(Z)$ for all $k \le K_a + 1$ and compute the ELBO after splitting, denoted by $\mathcal{L}_{CAVI}^{S}(\hat{\varpi}|Z)$.

---

(A4) If $\frac{\mathcal{L}_{CAVI}^{S}\left(\hat{\varpi}|Z\right)-\mathcal{L}_{CAVI}\left(\hat{\varpi}|Z\right)}{\mathcal{L}_{CAVI}\left(\hat{\varpi}|Z\right)}>\tau$, set $K_a = K_a + 1$, $\mathcal{L}_{CAVI}\left(\hat{\varpi}|Z\right)=\mathcal{L}_{CAVI}^{S}\left(\hat{\varpi}|Z\right)$, and go to step (A1), else break the loop.

**B. Greedy merge step:**

**Repeat:**

(B1) For each component $k \le K_a$, do:

- Compute $p\left(k_2|k_1\right)$ according to Eq. (16) for all $k_2 \le K_a$ and $k_2 \ne k_1$.
- Merge the pair $\left(k_1, k_2\right)$ with the maximum $p\left(k_2|k_1\right)$ into one component $k_{12}$.
- Compute the summary statistics $s_{k_{12}}\left(Z\right)$ and the new ELBO after merging $\mathcal{L}_{CAVI}^{M(k_1,k_2)}\left(\hat{\varpi}|Z\right)$, then revert to the configuration before merging.

(B2) Find the pair $\left(K_1, K_2\right)$ with $\mathcal{L}_{CAVI}^{M(K_1,K_2)}\left(\hat{\varpi}|Z\right)=\max\left\{\mathcal{L}_{CAVI}^{M(k_1,k_2)}\left(\hat{\varpi}|Z\right)\right\}_{k_1=1}^{K_a}$ and merge them into one component $K_{12}$.

(B3) Update the summary statistics for all components and compute the ELBO after merging, denoted by $\mathcal{L}_{CAVI}^{M}\left(\hat{\varpi}|Z\right)$.

(B4) If $\frac{\mathcal{L}_{CAVI}^{M}\left(\hat{\varpi}|Z\right)-\mathcal{L}_{CAVI}\left(\hat{\varpi}|Z\right)}{\mathcal{L}_{CAVI}\left(\hat{\varpi}|Z\right)}>\tau$, set $K_a = K_a - 1$, $\mathcal{L}_{CAVI}\left(\hat{\varpi}|Z\right)=\mathcal{L}_{CAVI}^{M}\left(\hat{\varpi}|Z\right)$, and go to step (B1), else break the loop.

Compute the variational posterior based on the prior and the summary statistics.

**Output:** Number of active components $K_a$; summary statistics $\left\{s_k\left(Z\right)\right\}_{k=1}^{K_a}$; optimized variational parameters $\hat{\varpi}=\left\{\hat{\alpha}_k, \hat{\varphi}_k, \hat{\pi}_{nk}\right\}_{k=1,2,\ldots,K_a}^{n=1,2,\ldots,N}$.

### 3.4 Training procedure of DPVIIL

Based on Section 3.2 and Section 3.3, the detailed procedures of optimizing DPVIIL can be summarized as follows:

**Algorithm 2:** DPVIIL

**Input:** Dataset $X=\left\{x_n\right\}_{n=1}^{N}$, batch size $B$, initial network parameters $\theta$ and $\phi$, initial parameters of the DPMM $K_a = 1$, $\varpi=\left\{\alpha,\varphi\right\}$, $s_k\left(Z\right)$.

**Output:** Optimized network parameters $\theta,\phi$ and DPMM parameters $K_a, \hat{\varpi}=\left\{\hat{\alpha}_k, \hat{\varphi}_k, \hat{\pi}_{nk}\right\}_{k=1,2,\ldots,K_a}^{n=1,2,\ldots,N}, \left\{s_k\left(Z\right)\right\}_{k=1}^{K_a}$.

1. **Repeat:**
2. Create an empty dataset $Z=\varnothing$ to store the latent representations of each minibatch.
3. For each minibatch, do:
4. Sample a batch of data $X^B=\left\{x_n\right\}_{n=1}^{B}$ from the dataset $X$.
5. Fix the DPMM parameters, compute the objective function for the neural networks $\mathcal{L}^B\left(\theta,\phi|\hat{\varpi},X^B\right)$.

Compute the gradients of minibatch estimator $\nabla_{\theta,\phi}\mathcal{L}^{B}\left(\theta,\phi \middle| \hat{\varpi}, X^{B}\right)$.
Update the parameters $\theta,\phi$ using the gradients.
Obtain the latent representations of current minibatch $Z^{B}=\{z_n\}_{n=1}^{B}$ through the reparameterization trick and store them into the dataset $Z=Z\bigcup Z^{B}$.
Fix the network parameters $\theta,\phi$ and update the parameters of the DPMM $K_a, \hat{\varpi}=\{\hat{\alpha}_k,\hat{\varphi}_k,\hat{\pi}_{nk}\}_{k=1,2,\ldots,K_a}^{n=1,2,\ldots,N}, \{s_k(Z)\}_{k=1}^{K_a}$ using the stored latent representations according to Algorithm 1.
Reset the dataset $Z=\varnothing$.
**Until Convergence**.

Algorithm 2 enables jointly optimizing the parameters of the neural networks and the DPMM through an iterative approach. Notably, as the entire optimization process alternates between updating the DPMM and the VAE, less rigorous convergence criteria can be used to improve the training efficiency of DPVIIL. Unlike the approach that uses a VAE for feature extraction followed by clustering with a DPMM, where the DPMM and VAE are trained independently, the joint optimization scheme of DPVIIL regularizes the learned latent space to align with a DPMM manifold [30], promoting it to extract discriminative features, dynamically adapt the number and shape of clusters, and effectively capture the underlying structure of the data during optimization. For incremental learning, information from previous data can be retained through the network parameters and the summary statistics of the DPMM, allowing DPVIIL to dynamically adapt to new data in few-shot or even zero-shot settings [36]. Furthermore, while DPVIIL is used as an illustrative example in this work, the DPGIIL framework can be generalized to other DGMs. For example, it can be implemented with flow-based DGMs [29] by introducing a normalizing flow in the latent space to transform the original latent representation $z$ into a variable $z_K$ with a more complex distribution.

In this case, the lower bound in Eq. (13) can be adapted accordingly using the method in [29]. This approach provides a more flexible latent space for representation learning but leads to increased computational costs.

For online structural anomaly detection, DPVIIL automatically learns informative representations from raw TF vectors and adaptively clusters them to identify distinct patterns of structural behavior influenced by EOVs or different damage scenarios. Additionally, the incremental learning capability of DPVIIL allows it to dynamically recognize and learn new patterns of structural behavior from incoming data during long-term operation, thus enabling timely identification of new structural conditions and supporting informed decision-making for detected anomalies.

# 4 Case Studies

The performance of DPVIIL is validated through three case studies: an 8-story numerical shear building and two benchmark datasets from the S101 bridge [44] and the Yonghe bridge [45]. The first two case studies include clearly defined healthy and distinct damaged states and are used to evaluate DPVIIL's incremental learning and clustering capabilities. In contrast, the Yonghe bridge dataset contains diverse healthy and damaged patterns influenced by EOVs but can only be coarsely labeled as "healthy" and "damaged". Therefore, this dataset is used to evaluate the anomaly detection performance of DPVIIL in real-world scenarios.

In these case studies, the inputs to DPVIIL are vectors containing the magnitudes of TFs over a wide frequency band encompassing multiple natural frequencies of the monitored structure, which are extracted from raw acceleration measurements between two adjacent sensors. The same model architecture of DPVIIL is used across all case studies, as detailed in Appendix III. Each experiment is repeated five times, and the average performance along with the standard deviation is recorded for analysis. Three performance metrics, including the damage detection accuracy (DDA), unsupervised clustering accuracy (ACC) [32], and normalized mutual information (NMI) [46], are employed, with higher values indicating better performance across all metrics. DDA assumes that the normal condition data in the training set are known a priori and evaluates the model's ability to detect structural anomalies by distinguishing clusters corresponding to normal conditions from those indicating structural damage, as illustrated in Section 2.3. ACC compares the unsupervised cluster assignments with the ground truth labels and then finds the best matching between them through the Hungarian algorithm [32], which provides an overall alignment between the cluster assignments and the ground truth labels. NMI measures the amount of shared information between inferred cluster assignments and the ground truth labels, which is normalized to account for varying cluster numbers and is independent of label permutation [46]. In the numerical and S101 case studies, ground truth class labels are added for performance evaluation but are not used during training.

## 4.1 Structure descriptions

### *4.1.1 Numerical shear building*

For the numerical structure, the stiffness and mass of each floor are set as $K_i = 2.5\times10^6\, N/m$ and $m_i = 1000kg$, respectively. Classical Rayleigh damping is applied, with the damping ratios for the first two modes assumed to be $\varsigma_1 = \varsigma_2 = 1\%$. In addition to the healthy condition, seven damage scenarios are simulated by reducing the stiffness of certain floors, as summarized in Table 1. The structure is excited by an ambient input at the first floor, which is modeled as Gaussian white noise with an auto-PSD of $0.5m^2s^{-3}$. The acceleration measurements are contaminated by Gaussian white noise with a signal-to-noise ratio of 20dB to simulate the effect of measurement noise. The sampling duration is set as 300s with a sampling frequency of 50Hz. A total of 5000 acceleration samples are generated to construct TF vectors, with the first 1500 from the healthy condition and the remaining 3500 from different damage scenarios.

Table 1. Damage scenarios for the 8-story numerical shear building.

| Scenario | Damaged floor | Damage extent | No. of samples | Class label |
|---|---|---|---|---|
| Healthy state | - | - | 1500 | 0 |
| Damage scenario 1 | 1 | 5% | 500 | 1 |
| Damage scenario 2 | 1 | 10% | 500 | 2 |
| Damage scenario 3 | 2<br>4 | 10%<br>10% | 500 | 3 |
| Damage scenario 4 | 1<br>3<br>5 | 10%<br>15%<br>20% | 500 | 4 |
| Damage scenario 5 | 2<br>4<br>6 | 15%<br>20%<br>25% | 500 | 5 |
| Damage scenario 6 | 1<br>3<br>5<br>7 | 10%<br>15%<br>20%<br>25% | 500 | 6 |
| Damage scenario 7 | 1<br>2<br>4<br>6<br>8 | 10%<br>15%<br>20%<br>25%<br>30% | 500 | 7 |

#### *4.1.2 S101 bridge*

The S101 bridge dataset contains acceleration measurements from a progressive damage test [44]. The sampling frequency of the sensors is 500Hz, with each measurement lasting 5.5 minutes. These 5.5-minute raw measurements are further divided into shorter segments with each lasting 150s to construct TF vectors, resulting in a total of 730 TF vector samples. The damage test artificially introduces two major damage scenarios [44]: pier lowering and tendon cutting. Between these two damage scenarios, repair work is conducted to simulate real-world bridge maintenance. Four class labels are manually assigned as ground truth to evaluate the clustering performance of DPVIIL, aiming to verify its capability to distinguish between different damage scenarios for real-world bridge structures. The structural conditions and their corresponding labels are summarized in Table 2. For more details on the S101 bridge and the progressive damage test, one can refer to [44].

Table 2. Structural states of the S101 bridge and their corresponding labels (reproduced from [44]).

| Structural State | Damage actions | | Corresponding effects | Class label |
|---|---|---|---|---|
| Initial healthy condition | A | No action | ♦ Baseline | 0 |
| Damage caused by pier lowering | B | Begin of cutting through column | ♦ No signs of extra cracking or increase of existing cracks | 1 |
| | C | End of second cut through the pier | ♦ Formation of an extra hinge just above the foundation, which itself is equivalent to a constructive fixed support | |
| | D | 1st step of the pier settlement (10mm) | ♦ Moderate noise | |
| | E | 2nd step of the pier settlement (20 mm) | ♦ Horizontal cracks are found in neighboring pier | |
| | F | 3rd step of the pier settlement (27 mm) | ♦ Settling of bridge deck until reaching the elastic limits, support is not lost completely due to the hydraulic jack | |
| Repair work | G | Inserting steel plates | ♦ No obvious effect | 2 |
| | H | Uplifting the damaged pier | ♦ Some cracks are closed<br>♦ Hinge from cutting remains | |
| Damage caused by tendon cutting | I | Exposing cables and cutting of 1st cable | ♦ Reduction of prestressing without indication of changing conditions | 3 |
| | J | Cutting through 2nd cable | ♦ No obvious influence on structural behavior since bridge is not loaded by traffic | |
| | K | Cutting through 3rd cable | ♦ No obvious effect | |
| | L | Partly cutting of 4th cable | ♦ The extra prestressing reservoir is run out | |

#### *4.1.3 Yonghe bridge*

The Yonghe bridge dataset [45] (https://www.cece.ucf.edu/IASCM/events/smc-benchmark-problems-for-condition-assessment-and-damage-detection) contains acceleration measurements collected during long-term monitoring of the bridge from January to August 2008. In August 2008, two damage patterns were detected: a closure segment crack at the side span and a detachment between the girder and auxiliary piers [45]. The sampling frequency of the sensors is 100Hz, with each measurement lasting 1 hour. From the raw acceleration measurements, 2139 TF vector samples are extracted, with the first 1714 corresponding to normal conditions and the remaining 425 to damaged states. As this dataset includes healthy patterns affected by EOVs along with real-world damage mechanisms, it has become a benchmark for evaluating the performance of structural anomaly detection methods in real-world scenarios [45, 47].

### 4.2 Evaluation of incremental learning and clustering performance

In this section, the incremental learning and clustering performance of DPVIIL is validated using the numerical and the S101 bridge datasets. Comparative studies with state-of-the-art clustering methods are also conducted to further verify the effectiveness of DPVIIL.

#### *4.2.1 Online structural anomaly detection based on incremental learning*

For online structural anomaly detection, DPVIIL is initially trained on a dataset containing only TF vectors from the normal condition. Subsequently, TF vectors from different damage scenarios are gradually observed and incorporated into the training process, which is a class-incremental problem in the context of incremental learning [48] and aligns with real-world structural anomaly detection scenarios. The ACC of DPVIIL across training epochs for both case studies is presented in Fig. 6. In the numerical case study, DPVIIL is initially trained on data from class 0 (the healthy state), with the number of classes (structural conditions) increased to 3, 5, and 8 at epochs 40, 80, and 190, respectively. For the S101 bridge case study, the model is first trained on data from class 0, and data from the remaining three structural states are incorporated at epochs 60, 120, and 180, respectively. As shown in Fig. 6, the dynamic adaptivity of the proposed method is clearly demonstrated. The ACC experiences sudden decreases at the training epochs when data from new structural conditions are introduced, while it quickly recovers to high levels as the training progresses, as the information learned previously is effectively retained in the summary statistics of the DPMM and the network parameters. It is worth mentioning that the fluctuations in ACC observed during the incremental learning process can be attributed to the use of a relaxed convergence criterion in each training epoch, as described in Section 3.4.

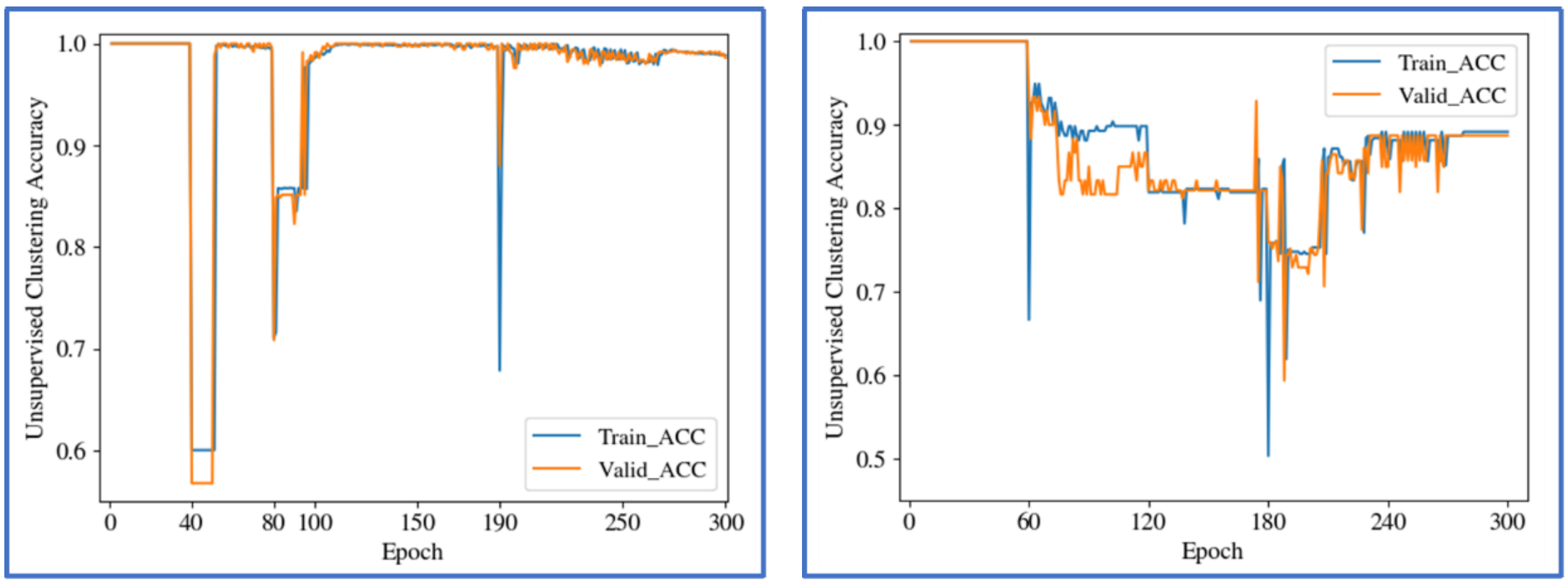

(a) (b)

Figure 6. The variation of ACC of DPVIIL across training epochs for (a) the numerical dataset and (b) the S101 bridge dataset.

To provide a more comprehensive evaluation of the incremental learning capability of DPVIIL, t-SNE [49] plots of the latent space of DPVIIL during training are presented in Fig. 7, while the performance metrics after observing all structural conditions are summarized in Table 3. From Fig. 7, one can find that compared to the ground truth labels, DPVIIL tends to assign samples that deviate from the center of a class to individual clusters with small sizes. This is reasonable as DPVIIL operates in a completely unsupervised manner here. For the numerical case study, all performance metrics exceed 0.95, demonstrating the sensitivity of the proposed method to changes in structural conditions. For the S101 bridge case study, the DDA remains at a high level (over 0.95), while the ACC and NMI exhibit a decrease. This can be caused by the repair work solely uplifting the damaged pier without addressing the cracks, making some vibration responses similar to those obtained at the onset of tendon cutting, as shown in Fig. 7b. These results illustrate the potential of DPVIIL for real-world online structural anomaly detection tasks, owing to its capability to continuously learn from incoming data without forgetting previously acquired knowledge. This allows DPVIIL to identify distinct normal and damaged structural states using clusters and to recognize new structural states by dynamically generating new clusters, making it well-suited for the long-term monitoring of engineering structures. Additionally, as DPVIIL can extract discriminative representations from raw data in an unsupervised manner, it can serve as a pre-training step for more complex techniques such as semi-supervised learning [19].

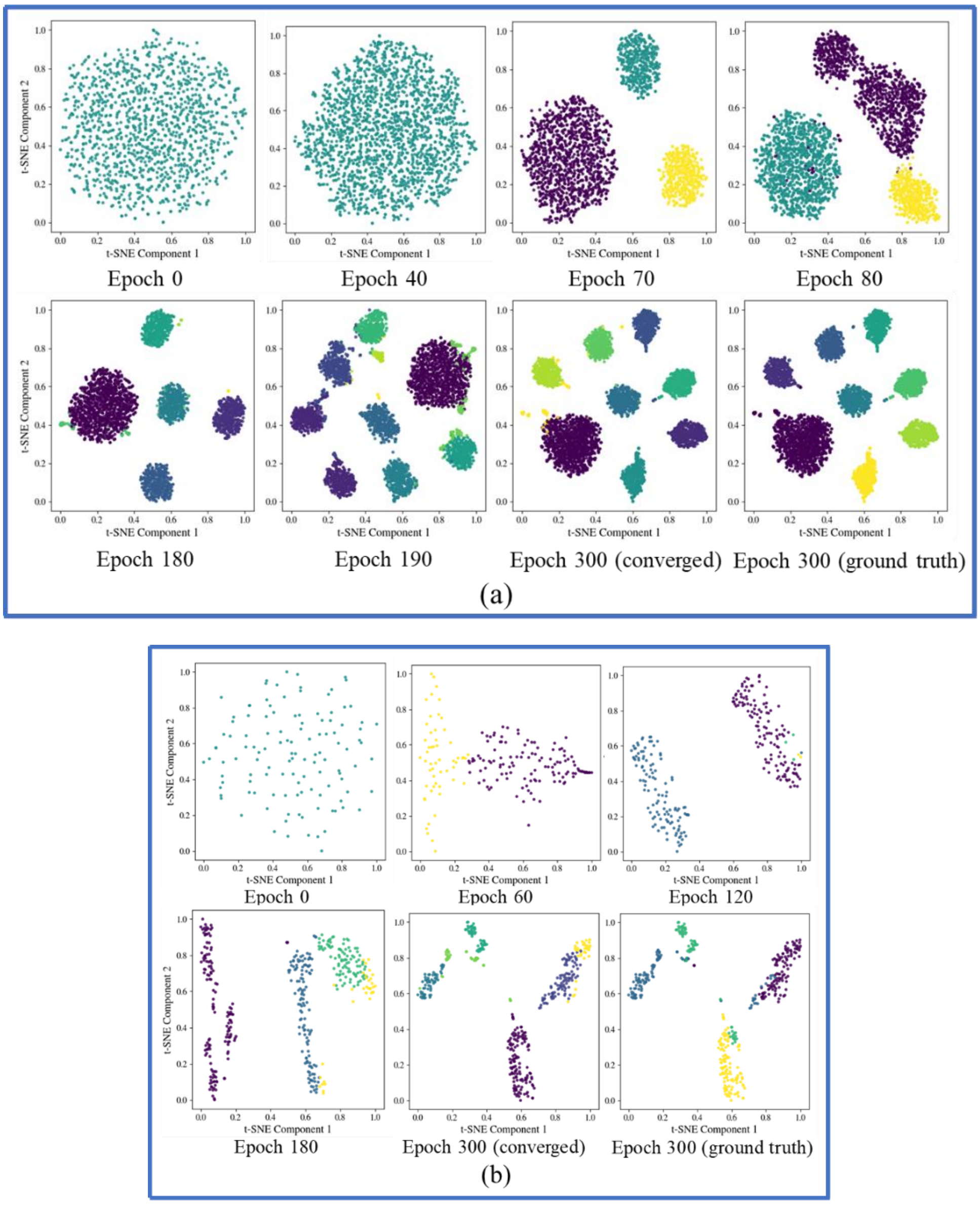

Figure 7. T-SNE plots of the latent space learned by DPVIIL during training for (a) the numerical dataset and (b) the S101 bridge dataset. Different colors represent either

cluster assignments or ground truth class labels.

Table 3. Performance of DPVIIL in the incremental learning scenario (mean±standard deviation of 5 runs).

| Dataset | Performance metrics | | |
|---|---|---|---|
| | DDA | ACC | NMI |
| Numerical shear building | 0.9932±0.0037 | 0.9920±0.0028 | 0.9809±0.0063 |
| S101 bridge | 0.9521±0.0171 | 0.8651±0.0232 | 0.8345±0.0264 |

#### *4.2.2 Comparative studies with state-of-the-art clustering algorithms*

To further analyze the performance of DPVIIL, it is compared with some state-of-the-art clustering algorithms in these two case studies. The results of the comparative studies are presented in Table 4, and the t-SNE plots of all compared approaches are shown in Fig. 8. The method "DPMM+VAE" refers to using a DPMM to cluster the latent space of a vanilla VAE, with the VAE and DPMM trained separately. This approach is included to highlight the importance of the proposed joint optimization scheme in regularizing the extracted latent representations. DEC (deep embedded clustering) [32], which requires the number of clusters to be predefined, is tested with 6, 8, and 10 clusters on the numerical dataset and with 3, 4, and 5 clusters on the S101 bridge dataset. This setup aims to evaluate the impact of the choice of cluster number on the performance of DEC and other clustering approaches that lack the ability to dynamically adjust the cluster number. For DIVA (Dirichlet Process Mixtures based Incremental deep clustering framework via Variational Auto-Encoder) [35], which also incorporates a DPMM, the concentration parameter is set as $\alpha = 10$ for both datasets, consistent with the configuration used for DPVIIL. Additionally, as DEC is designed for clustering in static scenarios and lacks the incremental learning capability [35], the comparative studies are conducted on complete datasets including all structural conditions. The VAE architecture for all compared approaches is identical to that used in DPVIIL, as detailed in Appendix III, while hyperparameters are tuned according to the original papers [32, 35].

From Table 4 and Fig. 8, one can find that for “DPMM+VAE”, the extracted latent representations are less discriminative compared to other methods, as the objective function of the vanilla VAE fails to adequately regularize the learned latent space [30], which highlights the advantage of jointly optimizing VAE and DPMM in the proposed method. For DEC, its performance is sensitive to the choice of the number of GMM components, with both over- and under-specification of the component number affecting its clustering performance. Additionally, the need for prespecifying the component number restricts its capability to dynamically adapt to new data for incremental learning, underscoring the advantages of employing a Bayesian nonparametric model. DIVA, on the other hand, tends to generate redundant clusters due to the relatively relaxed criterion for creating new components in the memoized online variational inference and its heuristic objective function. Compared to these approaches, DPVIIL introduces a DPMM prior into the latent space of the vanilla VAE to effectively regularize the extracted latent representations, which also endows DPVIIL with the dynamic adaptive capability for incremental learning by retaining previously learned information through summary statistics. Moreover, a reasonable optimization procedure is developed to enhance the model’s effectiveness and reliability. As a result, DPVIIL outperforms the compared methods in both structural anomaly detection and clustering performance across both case studies.

Table 4. Results of comparative studies on the numerical and S101 datasets (mean± standard deviation of 5 runs).

| Dataset | Numerical shear building | | | S101 bridge | | |
|---|---|---|---|---|---|---|
| Method | DDA | ACC | NMI | DDA | ACC | NMI |
| DPMM+VAE | 0.9504 ±0.0108 | 0.8244 ±0.0384 | 0.7757 ±0.0232 | 0.9452 ±0.0199 | 0.8466 ±0.0382 | 0.6827 ±0.0493 |
| DEC-6 (3) clusters | 0.9932 ±0.0048 | 0.8216 ±0.0183 | 0.9045 ±0.0091 | 0.9243 ±0.0051 | 0.6633 ±0.0069 | 0.4473 ±0.0349 |
| DEC-8 (4) clusters | 0.9888 ±0.0127 | 0.9208 ±0.0392 | 0.9505 ±0.0207 | 0.9113 ±0.0154 | 0.6942 ±0.0456 | 0.5033 ±0.0319 |
| DEC-10 (5) clusters | 0.9728 ±0.0102 | 0.9204 ±0.0393 | 0.9163 ±0.0206 | 0.9538 ±0.0266 | 0.7610 ±0.0249 | 0.6184 ±0.0188 |
| DIVA | 0.9856 ±0.0096 | 0.9884 ±0.0077 | 0.9745 ±0.0044 | 0.9627 ±0.0065 | 0.8151 ±0.0249 | 0.7569 ±0.0393 |
| **DPVIIL** | 0.9956 ±0.0050 | 0.9968 ±0.0010 | 0.9892 ±0.0066 | 0.9685 ±0.0201 | 0.8671 ±0.0455 | 0.8126 ±0.0341 |

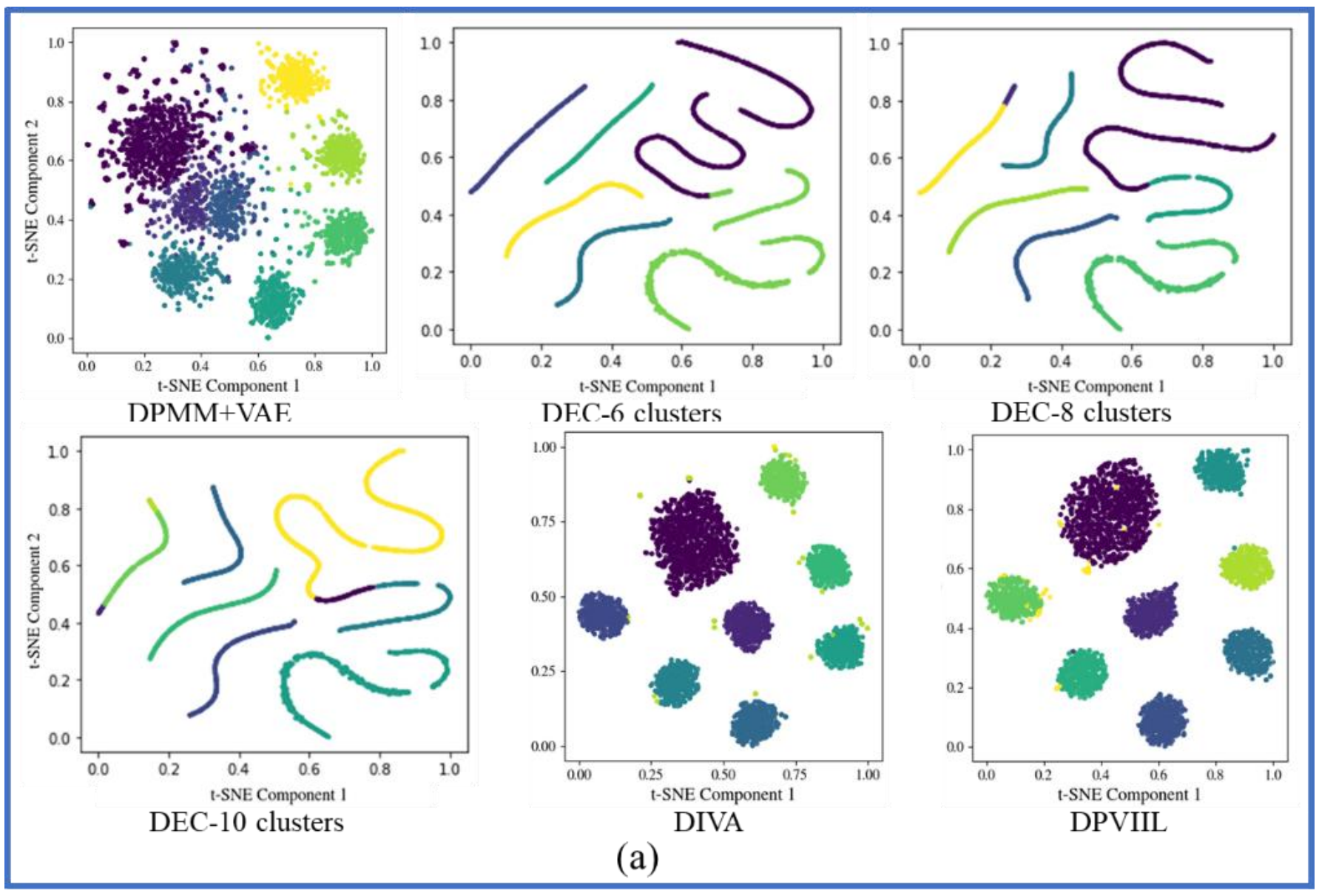

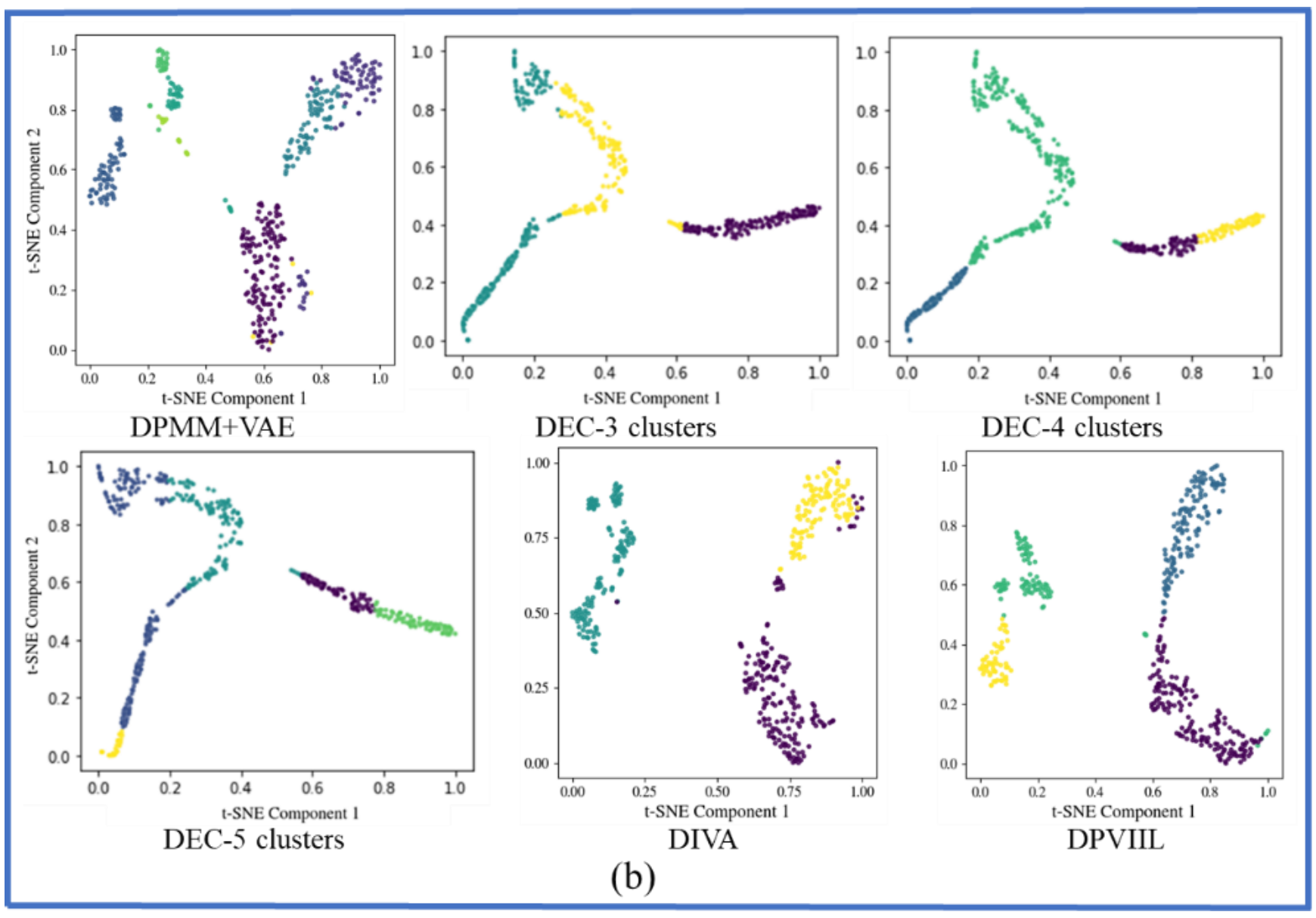

Figure 8. T-SNE plots of the latent space learned by the compared methods after convergence on (a) the numerical dataset and (b) the S101 bridge dataset. Different colors denote different cluster assignments.

### *4.2.3 Sensitivity analysis of hyperparameters*

The hyperparameters of DPVIIL include those related to the neural networks and the DPMM. The former can be tuned using some classical approaches [50], while the latter, especially the concentration parameter $\alpha$, is more challenging to adjust, as it directly affects the number of created clusters [19]. As a result, a sensitivity analysis is conducted to investigate the stability of DPVIIL with respect to $\alpha$, with the results provided in Table 5. From this table, it can be found that all performance metrics exhibit only slight variations in both case studies at different levels of $\alpha$. This can result from DPVIIL's iterative training process, where the DPMM regularizes the feature extraction by the neural networks, and the extracted latent representations, in turn, influence the update of DPMM parameters. As a result, the network parameters adapt automatically to different levels of $\alpha$ during training, leading to robust clustering performance that is less sensitive to changes in $\alpha$. This sensitivity analysis illustrates the robustness of the proposed method, thereby facilitating its practical implementation.

Table 5. Sensitivity analysis of DPVIIL with respect to the concentration parameter (mean±standard deviation of 5 runs).

| Dataset | Numerical shear building | | | S101 bridge | | |
|---|---|---|---|---|---|---|
| Alpha | DDA | ACC | NMI | DDA | ACC | NMI |
| 0.1 | 0.9952 ±0.0030 | 0.9952 ±0.0037 | 0.9859 ±0.0081 | 0.9713 ±0.0146 | 0.8849 ±0.0285 | 0.8484 ±0.0333 |
| 1 | 0.9960 ±0.0028 | 0.9968 ±0.0016 | 0.9901 ±0.0044 | 0.9617 ±0.0232 | 0.8808 ±0.0372 | 0.8202 ±0.0529 |
| 10 | 0.9956 ±0.0050 | 0.9968 ±0.0010 | 0.9892 ±0.0066 | 0.9685 ±0.0201 | 0.8671 ±0.0455 | 0.8126 ±0.0341 |
| 50 | 0.9956 ±0.0015 | 0.9948 ±0.0055 | 0.9853 ±0.0108 | 0.9576 ±0.0227 | 0.8918 ±0.0191 | 0.8262 ±0.0485 |
| 100 | 0.9972 ±0.0010 | 0.9968 ±0.0010 | 0.9909 ±0.0023 | 0.9534 ±0.0391 | 0.8822 ±0.0363 | 0.8157 ±0.0231 |

### 4.3 Evaluation of structural anomaly detection performance

As DPVIIL is applied to TF-based structural anomaly detection in this work, the Yonghe bridge dataset is used to further evaluate its performance in a real-world setting. To simulate a practical structural anomaly detection scenario, the first 1371 samples (80% of normal condition data) from the dataset are included in the training set to identify clusters representing normal operating conditions of the bridge, following the approach in [19]. During testing, any sample assigned to a cluster different from those representing the identified normal patterns is considered as an anomaly, indicating potential structural damage. In addition to DPVIIL, several state-of-the-art approaches are also tested on this dataset for comparison. These include anomaly detection methods such as the MSD-based method [5], the KNN-based method [51], one-class support vector machine (OCSVM) [51], and the VAE-based method [40], as well as the clustering methods evaluated in the previous two case studies, i.e., “DPMM+VAE”, DEC (with 4, 6, and 8 clusters), and DIVA. For the MSD-, KNN-, and VAE-based methods, the anomaly detection threshold is defined using the empirical quantile with a significance level of 5% [25]. For shallow learning methods, results from a single run are reported, whereas for DL approaches, each experiment is repeated five times with different random seeds, and the mean and standard deviation are reported.

Table 6 presents the results of the comparative study on the Yonghe dataset. It can be found that the evaluated DL methods generally outperform their shallow learning counterparts, owing to their ability to leverage deep neural networks for effective representation learning from raw TF vectors. Additionally, the tested clustering methods achieve better performance than the anomaly detectors. This can result from their use of mixture models to flexibly model the training data, allowing them to more effectively capture multiple healthy patterns under EOVs in this dataset. Meanwhile, their deep architectures further enhance their performance by facilitating the extraction of informative features from raw TF vectors. Apart from anomaly detection performance, clustering methods also provide more information about different patterns of structural behavior by assigning data to distinct clusters, which is an intrinsic advantage of clustering methods in structural anomaly detection and can facilitate more informed decision-making for the detected structural anomalies [20]. Furthermore, consistent with the results from the previous two case studies, DPVIIL achieves the best performance among the tested clustering methods, which can be attributed to its dynamic adaptivity and mathematically solid optimization framework. These results highlight the effectiveness and potential of DPVIIL for addressing real-world structural anomaly detection problems.

Table 6. Results of comparative studies on the Yonghe bridge dataset.

| Methods | DDA |
|---|---|
| MSD | 0.7784 |
| KNN | 0.8139 |
| OCSVM | 0.8467 |
| VAE | 0.8316±0.0051 |
| DPMM+VAE | 0.9230±0.0074 |
| DEC (4 clusters) | 0.8888±0.0569 |
| DEC (6 clusters) | 0.9511±0.0199 |
| DEC (8 clusters) | 0.9516±0.0209 |
| DIVA | 0.9747±0.0212 |
| **DPVIIL** | 0.9917±0.0107 |

## 5 Discussions

Structural anomaly detection is a critical task for ensuring the operational safety of engineering structures. By integrating a DPMM into the DGM framework, this work proposes DPGIIL, a deep clustering framework, for TF-based online structural anomaly detection. DPVIIL, a VAE-based implementation of DPGIIL, is used as an illustrative example and evaluated through three case studies. The results highlight its effectiveness and potential for real-world online structural anomaly detection tasks, as well as its superior performance compared to state-of-the-art methods. These advantages are attributed to its dynamic adaptability enabled by the Bayesian nonparametric clustering scheme and the mathematically solid framework for jointly optimizing the DPMM and network parameters. Nevertheless, there are still some practical challenges that warrant further investigation to enhance the robustness and applicability of this framework:

- Firstly, the structural anomaly detection performance of DPGIIL, as with other methods, depends on the quality of the training data from normal conditions, which are used to establish the baseline condition and identify the healthy patterns of the monitored structure. If the training data are insufficient, noisy, or unrepresentative, especially when influenced by EOVs, it would be challenging for DPGIIL to capture the full diversity of normal conditions, resulting in false negatives, false positives, and compromised anomaly detection accuracy. To mitigate this issue, future work can explore integrating data augmentation techniques, such as mixup [52] and GAN-based methods [53], or robustness testing strategies [54] into DPGIIL to improve its robustness when training data are limited or noisy.

- Secondly, DPGIIL integrates DGMs with DPMMs and relies on a complex optimization process integrating variational Bayesian inference with a greedy split-merge scheme. This sophisticated approach inevitably increases computational complexity compared to traditional methods, particularly when processing high-dimensional streaming data. In this study, all experiments are conducted on a workstation equipped with an NVIDIA RTX 4080 GPU and 32 GB of RAM. Such hardware requirements can pose challenges for real-world online structural anomaly detection systems with limited computing resources. To alleviate the computation burden, future work can explore parallel and distributed learning strategies [55] for DPGIIL to improve its efficiency and applicability.

## 6 Conclusions

This work proposes DPGIIL, a novel incremental clustering framework that combines DPMM and DGM for TF-based online structural anomaly detection, while DPVIIL serves as an illustrative example. By incorporating a DPMM prior into the latent space of a vanilla VAE, DPVIIL extracts discriminative features from raw TF vectors for both generative modeling and clustering, while the summary statistics of the DPMM, along with the network parameters, effectively retain information about previous data to enable incremental learning. To optimize DPVIIL, a tighter lower bound on its log marginal likelihood is derived, enabling jointly optimizing the network and DPMM parameters through a two-step iterative optimization approach. Additionally, a CAVI optimizer with a greedy split-merge scheme is devised to accelerate the optimization. Three case studies demonstrate that DPVIIL outperforms some state-of-the-art methods in structural anomaly detection and clustering. This is due to the adoption of a Bayesian nonparametric model to flexibly adapt model complexity to observed data, as well as a mathematically solid optimization framework. Furthermore, the incremental learning capability allows DPVIIL to dynamically generate new clusters as new data arrive to indicate the emergence of new structural conditions. It is worth mentioning that this framework can be flexibly extended to other DGMs and features for incremental clustering, while it also holds potential in data augmentation due to its generative modeling capability. These aspects warrant further investigation.

**Appendix I: Derivation of the tighter lower bound**

According to Eq. (5) and Eq. (9), the KL divergence term in the ELBO can be expressed as:

$$
\begin{aligned}
&D_{KL}\left[q\left(z,c,v,\eta \middle| x,\hat{\varpi},\phi\right) \middle\| p\left(z,c,v,\eta \middle| \varpi\right)\right] \\
&= \mathbb{E}_{q\left(z,c,v,\eta|x,\hat{\varpi},\phi\right)}\left[\log \frac{q\left(z|x,\phi\right)q\left(c,v,\eta \mid z,\hat{\varpi}\right)}{p\left(z \mid c,\eta\right)p\left(c,v,\eta|\varpi\right)}\right] \\
&= \mathbb{E}_{q\left(z|x,\phi\right)}\left[\log q\left(z|x,\phi\right)\right] - \mathbb{E}_{q\left(z,c,\eta|x,\hat{\varpi},\phi\right)}\left[\log p\left(z \mid c,\eta\right)\right] \\
&+ \mathbb{E}_{q\left(c,v,\eta|\hat{\varpi}\right)}\left[\log \frac{q\left(c,v,\eta \mid z,\hat{\varpi}\right)}{p\left(c,v,\eta \mid \varpi\right)}\right]
\end{aligned}
\tag{A1}
$$

As the DPMM is an infinite mixture model, the term $\mathbb{E}_{q\left(z,c,\eta|x,\hat{\varpi},\phi\right)}\left[p\left(z \mid c,\eta\right)\right]$ can be expressed as:

$$
\begin{aligned}
&\mathbb{E}_{q\left(z,c,\eta|x,\hat{\varpi},\phi\right)}\left[\log p\left(z \mid c,\eta\right)\right] \\
&= \int_{z} q\left(z|x,\phi\right)\sum_{k=1}^{\infty} q\left(c=k\right)\int_{\eta_k} q\left(\eta_k\right)\log p\left(z \mid \eta_k\right)d\eta_k dz \\
&= \sum_{k=1}^{\infty} q\left(c=k\right)\int_{z} q\left(z|x,\phi\right)\left[\int_{\eta_k} \mathcal{NW}\left(\eta_k \mid \hat{\varphi}_k\right)\log\mathcal{N}\left(z|\eta_k\right)d\eta_k\right]dz
\end{aligned}
\tag{A2}
$$

The integration $\int_{\eta_k} \mathcal{NW}\left(\eta_k \mid \hat{\varphi}_k\right)\log\mathcal{N}\left(z|\eta_k\right)d\eta_k$ is analytically intractable, and an intuitive approach is to approximate it through Monte Carlo integration given as follows:

$$
\begin{aligned}
&\mathbb{E}_{q\left(z,c,\eta|x,\hat{\varpi},\phi\right)}\left[\log p\left(z \mid c,\eta\right)\right] \\
&= \sum_{k=1}^{\infty} q\left(c=k\right)\int_{z_n} q\left(z|x,\phi\right)\left[\int_{\eta_k} \mathcal{NW}\left(\eta_k \mid \hat{\varphi}_k\right)\log\mathcal{N}\left(z|\eta_k\right)d\eta_k\right]dz \\
&\approx \sum_{k=1}^{\infty} q\left(c=k\right)\int_{z} q\left(z|x,\phi\right)\left[\frac{1}{J}\sum_{j=1}^{J}\log\mathcal{N}\left(z\middle|\eta_k^{(j)}\right)\right]dz,\ \eta_k^{(j)} \sim \mathcal{NW}\left(\eta_k \mid \hat{\varphi}_k\right) \\
&= \frac{1}{J}\sum_{j=1}^{J}\sum_{k=1}^{\infty} q\left(c=k\right)\int_{z} q\left(z|x,\phi\right)\log p\left(z\middle|\eta_k^{(j)}\right)dz
\end{aligned}
\tag{A3}
$$

As $\sum_{k=1}^{\infty} q\left(c=k\right)=1$, the KL divergence can be further simplified as:

$$
\begin{aligned}
&D_{KL}\left[q\left(z,c,v,\eta\middle|x,\hat{\varpi},\phi\right)\middle\|p\left(z,c,v,\eta\middle|\varpi\right)\right]\\
&=\mathbb{E}_{q(z|x,\phi)}\left[\log q\left(z\middle|x,\phi\right)\right]-\frac{1}{J}\sum_{j=1}^{J}\sum_{k=1}^{\infty}q\left(c=k\right)\int_{z}q\left(z\middle|x,\phi\right)\log p\left(z\middle|\eta_k^{(j)}\right)dz\\
&+\mathbb{E}_{q(c,v,\eta|z,\hat{\varpi})}\left[\log\frac{q\left(c,v,\eta\mid z,\hat{\varpi}\right)}{p\left(c,v,\eta\mid\varpi\right)}\right]\\
&=\frac{1}{J}\sum_{j=1}^{J}\sum_{k=1}^{\infty}q\left(c=k\right)\int_{z}q\left(z\middle|x,\phi\right)\log\frac{q\left(z\middle|x,\phi\right)}{p\left(z\middle|\eta_k^{(j)}\right)}dz+\mathbb{E}_{q(c,v,\eta|z,\hat{\varpi})}\left[\log\frac{q\left(c,v,\eta\mid z,\hat{\varpi}\right)}{p\left(c,v,\eta\mid\varpi\right)}\right]\\
&=\frac{1}{J}\sum_{j=1}^{J}\sum_{k=1}^{\infty}q\left(c=k\right)D_{KL}\left[q\left(z\middle|x,\phi\right)\middle\|p\left(z\middle|\eta_k^{(j)}\right)\right]+\mathbb{E}_{q(c,v,\eta|z,\hat{\varpi})}\left[\log\frac{q\left(c,v,\eta\mid z,\hat{\varpi}\right)}{p\left(c,v,\eta\mid\varpi\right)}\right]
\end{aligned}
\tag{A4}
$$

As $q\left(z\middle|x,\phi\right)=\mathcal{N}\left(z\middle|\mu_z,\sigma_z^2 I\right)$ and $p\left(z\middle|\eta_k^{(j)}\right)=\mathcal{N}\left(z\middle|\mu_k^{(j)},\left(\mathbf{\Lambda}_k^{(j)}\right)^{-1}\right)$, the KL divergence can be analytically derived. This approach provides an accurate approximation of the variational lower bound when the number of Monte Carlo samples $J$ is sufficiently large, but this comes at the cost of increased computational complexity. As a result, a more efficient approximation approach is proposed.

According to Jensen's inequality [56], it has:

$$
\begin{aligned}
&\mathbb{E}_{q(z,c,\eta|x,\hat{\varpi},\phi)}\left[\log p\left(z\mid c,\eta\right)\right]\\
&=\sum_{k=1}^{\infty}q\left(c=k\right)\int_{z}q\left(z\middle|x,\phi\right)\mathbb{E}_{q(\eta_k|\hat{\varphi}_k)}\left[\log p\left(z\middle|\eta_k\right)\right]dz\\
&\le\sum_{k=1}^{\infty}q\left(c=k\right)\int_{z}q\left(z\middle|x,\phi\right)\log\left(\mathbb{E}_{q(\eta_k|\hat{\varphi}_k)}\left[p\left(z\middle|\eta_k\right)\right]\right)dz\\
&=\sum_{k=1}^{\infty}q\left(c=k\right)\int_{z}q\left(z\middle|x,\phi\right)\log\left[t_{\hat{\upsilon}_k-D+1}\left(z\middle|\hat{m}_k,\frac{\left(\hat{\lambda}_k+1\right)\hat{\mathrm{W}}_k}{\hat{\lambda}_k\left(\hat{\upsilon}_k-D+1\right)}\right)\right]dz
\end{aligned}
\tag{A5}
$$

where $\hat{\varphi}_k=\left\{\hat{m}_k,\hat{\lambda}_k,\hat{\mathrm{W}}_k,\hat{\upsilon}_k\right\}$ denotes the parameters of the normal-Wishart distribution; $D$ is the dimensionality of $z$; $t_{\upsilon}(\bullet)$ is the Student's t distribution derived from the marginalization of the normal-Wishart distribution [57]. Therefore, it has:

$$
\begin{aligned}
& D_{KL}\left[q\left(z,c,v,\eta\middle|x,\hat{\varpi},\phi\right)\middle\|p\left(z,c,v,\eta\middle|\varpi\right)\right] \\
& \geq \mathbb{E}_{q(z|x,\phi)}\left[\log q\left(z\middle|x,\phi\right)\right]+\mathbb{E}_{q(c,v,\eta|\hat{\varpi})}\left[\log\frac{q\left(c,v,\eta\mid z,\hat{\varpi}\right)}{p\left(c,v,\eta\mid\varpi\right)}\right] \\
& -\sum_{k=1}^{\infty}q\left(c=k\right)\int_{z}q\left(z\middle|x,\phi\right)\log\left[t_{\hat{\upsilon}_k-D+1}\left(z\middle|\hat{m}_k,\frac{\left(\hat{\lambda}_k+1\right)\hat{\mathrm{W}}_k}{\hat{\lambda}_k\left(\hat{\upsilon}_k-D+1\right)}\right)\right]dz \\
& =\sum_{k=1}^{\infty}q\left(c=k\right)D_{KL}\left[q\left(z\middle|x,\phi\right)\middle\|t_{\hat{\upsilon}_k-D+1}\left(z\middle|\hat{m}_k,\frac{\left(\hat{\lambda}_k+1\right)\hat{\mathrm{W}}_k}{\hat{\lambda}_k\left(\hat{\upsilon}_k-D+1\right)}\right)\right] \\
& +\mathbb{E}_{q(c,v,\eta|\hat{\varpi})}\left[\log\frac{q\left(c,v,\eta\mid z,\hat{\varpi}\right)}{p\left(c,v,\eta\mid\varpi\right)}\right] \\
& \approx\sum_{k=1}^{\infty}q\left(c=k\right)D_{KL}\left[q\left(z\middle|x,\phi\right)\middle\|\mathcal{N}\left(z\middle|\hat{m}_k,\left(\hat{\lambda}_k\hat{\mathrm{W}}_k\right)^{-1}\right)\right]+\mathbb{E}_{q(c,v,\eta|\hat{\varpi})}\left[\log\frac{q\left(c,v,\eta\mid z,\hat{\varpi}\right)}{p\left(c,v,\eta\mid\varpi\right)}\right]
\end{aligned}
\tag{A6}
$$

Here a Gaussian distribution is used to approximate the Student's t distribution, with its mean and precision matrix set as the expected values of the normal-Wishart distribution, i.e. $\hat{m}_k$ and $\hat{\lambda}_k\hat{\mathrm{W}}_k$, thus making the KL divergence analytically tractable. The rationale of this approximation is that the degrees of freedom of the $k$ th Student's t distribution, $\hat{\upsilon}_k-D+1$, are proportional to the expected size of the $k$ th component of the DPMM [19], which makes the Student's t distribution approach a Gaussian distribution given suitable prior $\upsilon$ and component size [57].

According to the above analysis, the objective function becomes:

$$
\begin{aligned}
\mathcal{L}'\left(\theta,\phi,\hat{\varpi}\middle|x\right) &= \mathbb{E}_{q(z|x,\phi)}\left[\log p\left(x\middle|z,\theta\right)\right]-\mathbb{E}_{q(c,v,\eta|\hat{\varpi})}\left[\log\frac{q\left(c,v,\eta\mid z,\hat{\varpi}\right)}{p\left(c,v,\eta\mid\varpi\right)}\right] \\
& -\sum_{k=1}^{\infty}q\left(c=k\right)D_{KL}\left[q\left(z\middle|x,\phi\right)\middle\|\mathcal{N}\left(z\middle|\hat{m}_k,\left(\hat{\lambda}_k\hat{\mathrm{W}}_k\right)^{-1}\right)\right] \\
& \geq \mathbb{E}_{q(z|x,\phi)}\left[\log p\left(x\middle|z,\theta\right)\right]-D_{KL}\left[q\left(z,c,v,\eta\middle|x,\hat{\varpi},\phi\right)\middle\|p\left(z,c,v,\eta\middle|\varpi\right)\right] \\
& =\mathcal{L}_{ELBO}\left(\theta,\phi,\hat{\varpi}\middle|x\right)
\end{aligned}
\tag{A7}
$$

To make the objective a strictly tighter lower bound, the condition $\log p(x)\geq\mathcal{L}'\left(\theta,\phi,\hat{\varpi}\middle|x\right)$ must be satisfied. According to Jensen's inequality, it has:

$$\begin{aligned}
\log p(x) &= \log \int \frac{p(x,z,c,v,\eta|\varpi,\theta)}{q\left(z,c,v,\eta|x,\hat{\varpi},\phi\right)} q\left(z,c,v,\eta|x,\hat{\varpi},\phi\right) d\left(z,c,v,\eta\right) \\
&= \log \mathbb{E}_{q\left(z,c,v,\eta|x,\hat{\varpi},\phi\right)}\left[\frac{p(x|z,\theta)p\left(c,v,\eta|\varpi\right)}{q\left(z,c,v,\eta|x,\hat{\varpi},\phi\right)} p\left(z|c,\eta\right)\right] \\
&= \log\left(\mathbb{E}_{q\left(z,c,v,\eta|x,\hat{\varpi},\phi\right)}\left[\frac{p(x|z,\theta)p\left(c,v,\eta|\varpi\right)}{q\left(z,c,v,\eta|x,\hat{\varpi},\phi\right)}\right]\mathbb{E}_{q\left(z,c,v,\eta|x,\hat{\varpi},\phi\right)}\left[p\left(z|c,\eta\right)\right]+\right. \\
&\left.\operatorname{cov}\left(\frac{p(x|z,\theta)p\left(c,v,\eta|\varpi\right)}{q\left(z,c,v,\eta|x,\hat{\varpi},\phi\right)}, p\left(z|c,\eta\right)\right)\right)
\end{aligned} \tag{A8}$$

where $\operatorname{cov}(\bullet)$ is the covariance. If the covariance is positive or neglectable, it has:

$$\begin{aligned}
\log p(x) &\geq \log \mathbb{E}_{q\left(z,c,v,\eta|x,\hat{\varpi},\phi\right)}\left[\frac{p(x|z,\theta)p\left(c,v,\eta|\varpi\right)}{q\left(z,c,v,\eta|x,\hat{\varpi},\phi\right)}\right] + \log \mathbb{E}_{q\left(z,c,v,\eta|x,\hat{\varpi},\phi\right)}\left[p\left(z|c,\eta\right)\right] \\
&\approx \log \mathbb{E}_{q\left(z,c,v,\eta|x,\hat{\varpi},\phi\right)}\left[\frac{p(x|z,\theta)p\left(c,v,\eta|\varpi\right)}{q\left(z,c,v,\eta|x,\hat{\varpi},\phi\right)}\right] \\
&\quad + \log \sum_{k=1}^{\infty} q\left(c=k\right) \int_z q\left(z|x,\phi\right) \mathcal{N}\left(z \middle| \hat{m}_k, \left(\hat{\lambda}_k \hat{\mathrm{W}}_k\right)^{-1}\right) dz \\
&\geq \mathbb{E}_{q\left(z,c,v,\eta|x,\hat{\varpi},\phi\right)}\left[\log \frac{p(x|z,\theta)p\left(c,v,\eta|\varpi\right)}{q\left(z,c,v,\eta|x,\hat{\varpi},\phi\right)}\right] \\
&\quad + \sum_{k=1}^{\infty} q\left(c=k\right) \int_z q\left(z|x,\phi\right) \log \mathcal{N}\left(z \middle| \hat{m}_k, \left(\hat{\lambda}_k \hat{\mathrm{W}}_k\right)^{-1}\right) dz \\
&= \mathcal{L}'\left(\theta,\phi,\hat{\varpi}|x\right)
\end{aligned} \tag{A9}$$

Although the relationship between $\frac{p(x|z,\theta)p\left(c,v,\eta|\varpi\right)}{q\left(z,c,v,\eta|x,\hat{\varpi},\phi\right)}$ and $p\left(z|c,\eta\right)$ is complex and the sign of their covariance cannot be determined analytically, in practice, given a sufficiently large amount of training data, the influence of the likelihood on the posterior tends to dominant over the prior. This reduces the dependency between $\frac{p(x|z,\theta)p\left(c,v,\eta|\varpi\right)}{q\left(z,c,v,\eta|x,\hat{\varpi},\phi\right)}$ and $p\left(z|c,\eta\right)$, leading to a neglectable covariance. In this scenario, $\mathcal{L}'\left(\theta,\phi,\hat{\varpi}|x\right)$ can be regarded as a tighter bound to optimize DPVIIL. To make the derivation more rigorous and applicable to cases where training data are limited, a hyperparameter $\gamma$, which is set as $\gamma=1$ in our experiments, is introduced

to penalize the overestimate of the KL divergence in Eq. (A5) and ensure the objective function remains a valid lower bound on the log marginal likelihood. Consequently, the final lower bound becomes that given in Eq. (13).

**Appendix II: Update equations for the variational parameters of the DPMM using summary statistics**

For a DPMM with Gaussian components, the variational parameters to be optimized through CAVI are $\hat{\varpi}=\{\hat{\alpha},\hat{\varphi},\hat{\pi}\}$, as defined in Sections 2 and 3, where $\hat{\alpha}=\{\hat{\alpha}_{1,k},\hat{\alpha}_{2,k}\}_{k=1}^{K_a}$, $\hat{\varphi}=\{\hat{m}_k,\hat{\lambda}_k,\hat{\mathrm{W}}_k,\hat{\upsilon}_k\}_{k=1}^{K_a}$, and $\hat{\pi}=\{\hat{\pi}_{nk}\}$. Given a set of samples $Z=\{z_n\}_{n=1}^{N}$, the responsibility $\hat{\pi}_{nk}$, which denotes the probability of assigning $z_n$ to the $k$-th cluster, can be computed using the method in [19] given other variational parameters $\hat{\alpha}$ and $\hat{\varphi}$. Based on the responsibilities, the summary statistics for each component, i.e., $N_k=\sum_{n=1}^{N}\hat{\pi}_{nk}$, $\bar{z}_k=\frac{1}{N_k}\sum_{n=1}^{N}\hat{\pi}_{nk}z_n$, and $\mathrm{S}_k=\frac{1}{N_k}\sum_{n=1}^{N}\hat{\pi}_{nk}\left(z_n-\bar{z}_k\right)\left(z_n-\bar{z}_k\right)^T$, can be derived to update the variational parameters $\hat{\alpha}$ and $\hat{\varphi}$:

$$\begin{aligned}
&\hat{\alpha}_{1,k}=1+N_k;\ \hat{\alpha}_{2,k}=\alpha+N_k \\
&\hat{\lambda}_k=\lambda+N_k;\ \hat{m}_k=\frac{1}{\hat{\lambda}_k}\left(\lambda m+N_k\bar{z}_k\right);\ \hat{\upsilon}_k=\upsilon+N_k \\
&\hat{\mathrm{W}}_k^{-1}=\mathrm{W}^{-1}+N_k\mathrm{S}_k+\frac{\lambda N_k}{\lambda+N_k}\left(\bar{z}_k-m\right)\left(\bar{z}_k-m\right)^T
\end{aligned} \tag{B1}$$

where $\varpi=\{\alpha,\varphi\}=\{\alpha,m,\lambda,\mathrm{W},\upsilon\}$ denotes the prior parameters. Given these equations, the CAVI algorithm iteratively updates the responsibilities, summary statistics, and variational parameters $\hat{\alpha}$ and $\hat{\varphi}$. After these updates, the ELBO $\mathcal{L}_{CAVI}\left(\hat{\varpi}\middle|Z\right)=\mathbb{E}_{q(c,v,\eta|\hat{\varpi})}\left[\log\frac{p\left(Z,c,v,\eta\mid\varpi\right)}{q\left(c,v,\eta\mid Z,\hat{\varpi}\right)}\right]$ is computed using the formula in [19]. This process repeats until the ELBO converges, yielding an optimized variational posterior for the DPMM. For incremental learning, when a new dataset $Z' = \{z_{n'}\}_{n'=1}^{N'}$ is obtained, the summary statistics can be directly updated without reanalyzing previous data:

$$\begin{aligned} N_k^{new} &= N_k + N_k' \\ \bar{z}_k^{new} &= \frac{N_k \bar{z}_k + N_k' \bar{z}_k'}{N_k + N_k'} \\ \mathrm{S}_k^{new} &= \mathrm{S}_k + \mathrm{S}_k' + \frac{N_k N_k'}{N_k + N_k'} \left(\bar{z}_k - \bar{z}_k'\right)\left(\bar{z}_k - \bar{z}_k'\right)^T \end{aligned} \tag{B2}$$

where $N_k' = \sum_{n'=1}^{N'} \hat{\pi}_{n'k}$, $\bar{z}_k' = \frac{1}{N_k'} \sum_{n'=1}^{N'} \hat{\pi}_{n'k} z_{n'}$, and $\mathrm{S}_k' = \frac{1}{N_k'} \sum_{n'=1}^{N'} \hat{\pi}_{n'k} \left(z_{n'} - \bar{z}_k'\right)\left(z_{n'} - \bar{z}_k'\right)^T$. These updated summary statistics can then be used to optimize the variational parameters.

## Appendix III: Implementation details of DPVIIL in the case studies

For both case studies, we use the same VAE architecture to form DPVIIL, which is composed of fully connected (FC) layers with rectified linear unit (ReLU) activations, and the linear outputs of decoder are directly used as reconstructions without activation node. The structure is presented in Fig. A1, while the key hyperparameters involved in DPVIIL are listed in Table A1.

Table A1. General key hyperparameters of DPVIIL used in the case studies.

| Dataset | Numerical | S101 bridge | Yonghe bridge |
|---|---|---|---|
| Latent dim | 10 | 10 | 20 |
| Optimizer | Adam | Adam | Adam |
| Learning Rate | 5e-5 | 1e-5 | 5e-6 |
| Weight Decay | 0 | 0 | 0 |
| Train Epochs | 150 | 180 | 100 |
| Batch Size | 32 | 16 | 32 |
| Concentration parameter | 10 | 10 | 10 |
| Penalization factor $\gamma$ | 1 | 1 | 1 |
| Train/Validation/Test Split | 0.8/0.1/0.1 | 0.6/0.2/0.2 | - |

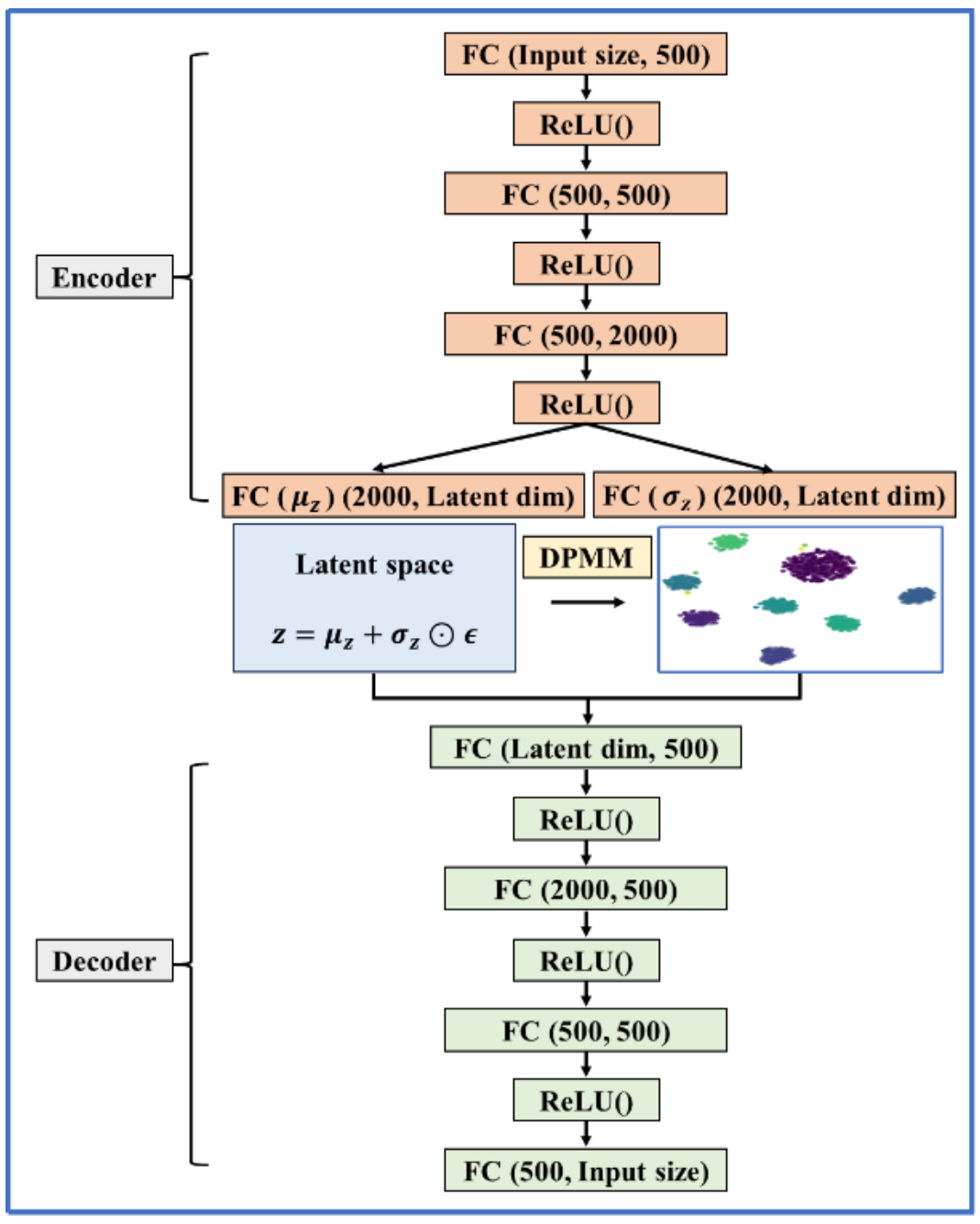

Figure A1. The architecture of DPVIIL used in the case studies.

**Acknowledgements:**

This research has been supported by the Science and Technology Development Fund, Macau SAR (File no.: 101/2021/A2, 0038/2024/RIB1, 001/SKL/2024), the Research Committee of University of Macau (File no.: MYRG-GRG2025-00270-IOTSC and MYRG-GRG2024-00119-IOTSC), Guangdong-Hong Kong-Macau Joint Laboratory Program (Project No.: 2020B1212030009). Also, the authors highly appreciate VCE for sharing the field test data of S101 Bridge.

**Declaration of generative AI and AI-assisted technologies in the writing process**

During the preparation of this work the authors used ChatGPT in order to improve the language. After using this tool, the authors reviewed and edited the content as needed and take full responsibility for the content of the publication.